\documentclass{article}

     \PassOptionsToPackage{numbers, compress}{natbib}
\usepackage[preprint]{neurips_2021}

\usepackage[utf8]{inputenc} % allow utf-8 input
\usepackage[T1]{fontenc}    % use 8-bit T1 fonts
\usepackage{hyperref}       % hyperlinks
\usepackage{url}            % simple URL typesetting
\usepackage{booktabs}       % professional-quality tables
\usepackage{amsfonts}       % blackboard math symbols
\usepackage{nicefrac}       % compact symbols for 1/2, etc.
\usepackage{microtype}      % microtypography
\usepackage[dvipsnames]{xcolor}
\usepackage{microtype}
\usepackage{graphicx}
\usepackage{booktabs} % for professional tables
\usepackage{subcaption}
\usepackage{multirow}
\usepackage{algorithm}
\usepackage{algorithmic}

\usepackage{mathtools}
\usepackage{amsmath,amsthm,amssymb,amsfonts}

\usepackage{comment}
\newtheorem{definition}{Definition}

\colorlet{Mycolor1}{green!10!orange!90!}

\title{SARC: Soft Actor Retrospective Critic}

\author{%
  Sukriti Verma\thanks{Work done while at Adobe} \\
  Carnegie Mellon University\\
  \texttt{sukritiv@andrew.cmu.edu} \\
   \And
   Ayush Chopra \\
   MIT Media Lab \\
  \texttt{ayushc@mit.edu} \\
   \AND
   Jayakumar Subramanian \\
   Adobe \\
  \texttt{jasubram@adobe.com} \\
   \And
  Mausoom Sarkar \\
   Adobe \\
   \texttt{msarkar@adobe.com} \\
  \And
   Nikaash Puri \\
   Adobe \\
   \texttt{nikpuri@adobe.com} \\
\And
   Piyush Gupta \\
   Adobe \\
   \texttt{piygupta@adobe.com} \\
   \And
   Balaji Krishnamurthy \\
   Adobe \\
   \texttt{kbalaji@adobe.com} \\
}

\begin{document}

\maketitle

\begin{abstract}
The two-time scale nature of SAC, which is an actor-critic algorithm, is characterised by the fact that the critic estimate has not converged for the actor at any given time, but since the critic learns faster than the actor, it ensures eventual consistency between the two. Various strategies have been introduced in literature to learn better gradient estimates to help achieve better convergence. Since gradient estimates depend upon the critic, we posit that improving the critic can provide a better gradient estimate for the actor at each time. Utilizing this, we propose Soft Actor Retrospective Critic (SARC),  where we augment the SAC critic loss with another loss term - retrospective loss - leading to faster critic convergence and consequently, better policy gradient estimates for the actor. An existing implementation of SAC can be easily adapted to SARC with minimal modifications. Through extensive experimentation and analysis, we show that SARC provides consistent improvement over SAC on benchmark environments. We plan to open-source the code and all experiment data at \url{https://github.com/sukritiverma1996/SARC}. 

%To provide some intuition, we also interpret the retrospective loss, a non-stationary adaptive objective function, as a contraction mapping which we derive as an equivalence of a standard fixed point theorem. 
\end{abstract}

\section{Introduction}

% [-replicate succeses of DL/supervised techniques in RL
% -scaling in RL achieved by bringing DL in RL
% -there have been advances of use of DL in RL
% -AC methods, success - policy based part  bring greater control, value based  - difference between reinforce and AC is that AC pays attention to structure
% -two diff interwoven loops - policy improvement - policy grad and policy eval (critic update)
% -advances in AC - A3C, TRPO, PPO, DPG, TD3, SAC

% -brief on policy grad theorem
% -now this is only stochastic grad descent with a moving target
% -given a critic, policy grad determines policy improvement
% -given policy, q function unique
% -td3, sac - only 2 algos that focus on critic - two q's and entropy - maximization bias in q (double q learning paper Sutton)
% -so what more can be done for the critic

% -Retrospective Loss
% -does not change global minimum

% -Marrying the two

% Claims
% -identifying the connection between retrospection and contraction thm
% -connect this to the two time-scale AC algos
%     -better q values, better grad, in turn better future samples
% -Using this theory to create SARC
%     -SAC is SOTA in rl
%     -extensive empirical demonstration and analysis
%     -better returns, no degrading, minimal edit to orig SAC to implement]

The idea of applying and extending successful supervised learning techniques from the field of deep learning to the field of reinforcement learning has long existed and helped achieve remarkable success in deep RL over this decade. 
One of the early breakthroughs, Deep Q-Networks (DQN) \cite{mnih2013playing, mnih2015human}, leveraged advances from the field of deep learning to train an end-to-end deep reinforcement learning agent, which achieved better than human-level performance in several challenging Atari games. 
Before DQNs, reinforcement learning could only be applied to tasks having low-dimensional state spaces or tasks where a small number of features could be extracted. 
With DQNs, it became possible to learn policies directly from high-dimensional input. 
This subsequently gave rise to a plethora of deep RL algorithms that combined deep learning with reinforcement learning. 
Broadly, these algorithms fall into three classes - critic-only (like DQN, DDQN \cite{van2016deep} etc.), actor-only (REINFORCE \cite{williams1992simple}) and actor-critic (SAC \cite{haarnoja2018soft}, PPO \cite{schulman2017proximal} etc.) algorithms, In this work, we seek to improve the Soft Actor-Critic Algorithm (SAC) \cite{haarnoja2018soft}, which is one of the current state-of-the-art actor-critic algorithms in Deep RL, by taking inspiration from a recent advance in the field of supervised deep learning - Retrospective Loss.

Actor-critic methods such as PPO, A3C \cite{mnih2016asynchronous}, DPG \cite{silver2014deterministic}, TD3 \cite{fujimoto2018addressing} and SAC have yielded great success in recent times \cite{wang2016sample, bahdanau2016actor}. With actor-critic algorithms, the central idea is to split the learning into two key modules: i) \textit{Critic}, which learns the policy-value function given the actor's policy. ii) \textit{Actor}, which uses the critic's policy-value function to estimate a policy gradient and improve its policy.
% \begin{enumerate}
%     \item Critic: The Critic learns the policy-value function given the Actor's policy.
%     \item Actor: The Actor improves its policy using stochastic gradient ascent, by estimating the policy gradient based on the policy-value function provided by the critic.
% \end{enumerate}

One iteration of policy improvement in actor-critic methods \cite{konda2000actor} involves two interwoven steps: one step of critic learning followed by one step of actor learning. These steps are interwoven for the following reason. Given a fixed critic, policy improvement and actor learning is determined by policy gradient estimates computed using the policy-value function provided by the critic. Given a fixed actor, the policy remains unchanged for the critic and hence, there exists a unique value function for the critic to learn. This allows the application of policy evaluation algorithms to learn this value function. A good value function, in turn enables accurate policy gradient estimates for the actor \cite{sutton2018reinforcement, wu2020finite}.

%After many steps of the Critic update, a good estimation of the value function can be expected.  A more favorable implementation is the so-called two time-scale actor-critic algorithm, where the actor and the critic are updated simultaneously at each iteration except that the actor changes more slowly than the critic. In this way, it can be expected that the Critic will be well approximated even after one iteration of policy improvement. 

Several advances in actor-critic methods such as TRPO \cite{schulman2015trust}, PPO, A3C and DDPG \cite{lillicrap2015continuous} have come from improving actor learning and discovering more stable ways of updating the policy. A few actor-critic algorithms, such as TD3 and SAC, have advanced state-of-the-art by improving critic learning. Both of these algorithms learn two value functions rather than one to counter the overestimation bias \cite{thrun1993issues} in value learning. TD3 updates the critic more frequently than the actor to minimize the error in the value function \cite{fujimoto2018addressing}. SAC uses the entropy of the policy as a regulariser \cite{haarnoja2018soft}. In this work, we propose to further improve critic learning in SAC by applying a regulariser that accelerates critic convergence. 

This regulariser is inspired from a recent technique that improves performance in the supervised learning setting called retrospective loss \cite{jandial2020retrospective}. When retrospective loss is minimized along with the task-specific loss, the parameters at the current training step are guided towards the optimal parameters while being pulled away from the parameters at a previous training step.  In the supervised setting, retrospective loss results in better test accuracy.

%We reinterpret this retrospective loss in terms of the contraction property and show that this interpretation implies convergence can be accelerated using this loss. 
Using this loss as a retrospective regulariser applied to the critic, accelerates the convergence of the critic. Due to the two time-scale nature of actor-critic algorithms \cite{borkar1997actor, konda2000actor}, causing the critic to learn faster leads to better policy gradient estimates for the actor, as gradient estimates depend on the value function learned by the critic. Better policy gradient estimates for the actor in turn lead to better future samples. 

Bringing it all together, in this work, we propose a novel actor-critic method, Soft Actor Retrospective Critic (SARC), which improves the existing Soft Actor-Critic (SAC) method by applying the aforementioned retrospective regulariser on the critic. We perform extensive empirical demonstration and analysis to show that SARC leads to better actors in terms of sample complexity as well as final return in several standard benchmark RL tasks provided in the DeepMind Control Suite and PyBullet Environments. We also show how SARC does not degrade performance for tasks where SAC already achieves optimal performance. To help reproducibility and to help convey that it takes minimal editing and essentially no added computation cost to convert an existing implementation of SAC to SARC, we open-source the code and all experiment data at \url{https://github.com/sukritiverma1996/SARC}. 

%\textcolor{blue}{We also reinterpret this retrospective loss in terms of the contraction property and show that this interpretation implies convergence can be accelerated using this loss in the appendix.}

\section{Related Work}\label{section:related-work}
Out of the three broad classes of RL algorithms, critic-only methods can efficiently learn an optimal policy \cite{jin2018q} (implicitly derived from the optimal action value function) under tabular setting. But under continuous or stochastic settings, function approximation can cause critic-only methods to diverge \cite{wiering2004convergence}. Actor-only methods can be applied to these settings but they suffer from high variance when computing the policy gradient estimates. Actor-critic algorithms combine the advantages of critic-only methods and actor-only methods. The key difference between actor-critic and actor-only algorithms is that actor-critic algorithms estimate an explicit critic or policy-value function instead of the Monte Carlo return estimated by the actor-only algorithms. In the sequel, we restrict attention to actor-critic algorithms. A more detailed comparison can be found in \citet{sutton2018reinforcement}.

Methods that model an explicit policy, such as actor-only and actor-critic algorithms, consist of two steps. i) \textit{Policy evaluation}: Calculating the policy-value function for the current policy, and ii) \textit{Policy improvement}: Using the value function to learn a better policy.

It is customary in RL to run these steps simultaneously utilizing the two time-scale stochastic approximation algorithm \cite{borkar1997actor,sutton1999policy,konda2000actor} rather than running these steps to convergence individually. This is the approach used in actor-critic algorithms, where the critic learns the policy-value function given the actor's policy and the actor improves it's policy guided by policy gradient estimates computed using the policy-value function provided by the critic.

%Many actor-critic algorithms build on the standard, on-policy policy gradient formulation to update the actor (Peters & Schaal, 2008), and many of them also consider the entropy of the policy, but instead of maximizing the entropy, they use it as an regularizer (Schulman et al., 2017b; 2015; Mnih et al., 2016; Gruslys et al., 2017). 

RL algorithms can be on-policy or off-policy. Off-policy actor-critic algorithms like Deep Deterministic Policy Gradient (DDPG) \cite{lillicrap2015continuous} and Twin Delayed DDPG (TD3) \cite{fujimoto2018addressing} often have a better sample complexity than on-policy algorithms. DDPG is a deep variant of the deterministic policy gradient algorithm \cite{silver2014deterministic}. It combines the actor-critic approach with insights from DQN. 
%and can be viewed both as a deterministic Actor-Critic algorithm and an approximate Q-learning algorithm. 
%DDPG uses a Q-function estimator to enable off-policy learning, and a deterministic actor that maximizes this Q-function. 
However, DDPG suffers from an overestimation bias \cite{thrun1993issues} in learning the Q-values. This hinders policy improvement. TD3 addresses this by learning two Q-functions and using the smaller of the two to compute the policy gradient. This favors underestimations. TD3 also proposes delaying policy updates until the value estimates have converged.

%There have been efforts to increase the sample efficiency while retaining robustness by incorporating off-policy samples and by using higher order variance reduction techniques (O’Donoghue et al., 2016; Gu et al., 2016). 

On-policy algorithms provide better stability but are not as sample efficient and often employ replay buffers to improve sample efficiency. Trust Region Policy Optimization (TRPO) \cite{schulman2015trust} proposes updating the policy by optimizing a surrogate objective, constrained on the KL-divergence between the previous policy and the updated policy. Actor Critic using Kronecker-Factored Trust Region (ACKTR) \cite{wu2017scalable} uses kronecker-factored approximated curvature to perform a similar trust region update. Proximal Policy Optimization (PPO) \cite{schulman2017proximal} proposes to match the performance of TRPO with only first-order optimization. 

Soft Actor-Critic (SAC) \cite{haarnoja2018soft} is an off-policy algorithm that has both the qualities of sample efficiency and stability. SAC optimizes a stochastic policy with entropy regularization. The policy is trained to maximize a trade-off between expected return and entropy. Increasing entropy results in more exploration, which can help accelerate learning. It can also prevent the policy from converging to a local optimum.

\paragraph{Our contribution}In this work, we propose Soft Actor Restrospective Critic (SARC), an algorithm that improves SAC by applying a retrospective regulariser on the critic inspired from retrospective loss \cite{jandial2020retrospective}. %We demonstrate the theoretical intuition behind the utility of this technique for improving SAC. 
We empirically validate this claim by comparing and analysing SARC with SAC, TD3 and DDPG on multiple continuous control environments.

\section{Soft Actor Retrospective Critic}

% a) Premise:

% b) Faster Critics and Better Actors
%     - Conjecture

% c) Theory: Retrospective Loss and Acceleration
%     - Definition and Theorem
%     - Validation (GLM and MLP)
    
% c) 
    
% d) Formalising SARC
%     - Algorithmic Block
%     - Policy Gradient Update Equation

%    - Discussion: Retrospective Loss for Moving Targets
    %     - Empirical Discussion {Fig 2 and 3}
    
%     NOTE: Minimal Edits + No hyperparam tuning.

\subsection{Premise}
In actor-critic methods, we note that the actor learning algorithm incrementally improves its policy, using the critic's performance estimate, while the best performance is unknown. However, in contrast to this, the critic learning algorithm is a supervised learning one albeit with a moving target. This is the key observation that enables us to import retrospective loss from supervised learning to actor-critic methods. 
%  There are two sources of non-stationarity in the ground truth---initialization of the Critic (bootstrapping effect) and non-stationarity of Actor's policy (accounted for in the two time-scale stochastic approximation approach).
% \subsection{Faster Critics and Better Actors} 

\paragraph{Retrospective loss and critic learning } We begin by describing our notation. Let $(x^{i}, y^{i})_{i=1}^{n_d} \in \mathcal{D}$ denote an input and output (ground truth) training data point, with  $n_d = |\mathcal{D}|$ denoting the size of the training data and $x^{i}, y^{i}$ belonging to the spaces $\mathcal{X} \subset \mathbb{R}^{n_x}$ and $\mathcal{Y} \subset \mathbb{R}^{n_y}$ respectively, where $n_x, n_y \in \mathbb{N}$ are the respective dimensions of the input and output. Furthermore, let $G_{\theta}: \mathcal{X} \to \mathcal{Y}$ denote a mapping from the input space to the output space, parameterized by $\theta \in \mathbb{R}^{n_{\theta}}$, where $n_{\theta} \in \mathbb{N}$ denotes the number of parameters in this mapping. Let $\hat{y}^{i}_{\theta} = G_{\theta}(x^{i})$ denote the predicted output for the data point $x^{i}$ using parameter $\theta$. The supervised learning objective is to learn a parameter $\theta^*$, which minimizes the error between the ground truth output $y^{i}$ and the predicted output $\hat{y}^{i}_{\theta^*}$ for all $i \in \{1, \dots, n_d \}$. Various metrics can be used to represent this loss, but for our illustration, we choose a loss based on the $L^2$ metric. We call this loss as the original loss, $\mathcal{L}_{\texttt{ori}}(\theta)$, which is defined as:
\begin{equation} \label{eq:method-loss-ori}
    \mathcal{L}_{\texttt{ori}}(\theta) = \frac{1}{n_d}\sum_{i=1}^{n_d}{(y^{i} - \hat{y}^{i}_{\theta})}^2
\end{equation}

We paraphrase the definition of retrospective loss from \citep{jandial2020retrospective}, and modify it slightly as follows:
\begin{definition} \label{def:retLoss}
[adapted from \citep{jandial2020retrospective}] Retrospective loss is defined in the context of an iterative parameter estimation scheme and is a function of the parameter values at two different iterations (times)---the current iteration and a previous iteration. Let $t, p$ denote the current and previous times respectively, and we so have $p < t$. Then, the retrospective loss at time $t$, is comprised of two components - the original loss and the retrospective regularizer as given below:
\begin{align}\label{eq:eq:method-loss-tot}
    \mathcal{L}_{\texttt{tot}}(\theta_t, \theta_p,(x^{i}, y^{i})_{i=1}^{n_d}) &= \mathcal{L}_{\texttt{ori}}(\theta_t, (x^{i}, y^{i})_{i=1}^{n_d}) \notag \\ 
    & + \mathcal{L}_{\texttt{ret}}(\theta_t, \theta_p, (x^{i}, y^{i})_{i=1}^{n_d}),
\end{align}
where the retrospective regularizer $\mathcal{L}_{\texttt{ret}}$ is defined as:
\begin{align}\label{eq:eq:method-loss-ret}
    &\mathcal{L}_{\texttt{ret}}(\theta_t, \theta_p, (x^{i}, 
    y^{i})_{i=1}^{n_d}) = \notag \\
    & \frac{1}{n_d}\sum_{i=1}^{n_d}\Bigl(
    (\kappa+1)\textup{d}_{\texttt{ret}}(y^{i}, \hat{y}^{i}_{\theta_t}) 
    - \kappa
    \textup{d}_{\texttt{ret}}(\hat{y}^{i}_{\theta_t},
    \hat{y}^{i}_{\theta_p})\Bigr),
\end{align}
where $\kappa > 0$ and $\textup{d}_{\texttt{ret}}(\cdot, \cdot)$ is any metric on vector spaces, such as $L^1$. We term the family of functions as given in~\eqref{eq:eq:method-loss-tot}, defined by parameters $\kappa$ and the metric $\textup{d}_{\texttt{ret}}(\cdot, \cdot)$ as retrospective loss functions.
\end{definition}

When retrospective loss is minimized along with the task-specific loss, the parameters at the current training step are guided towards the optimal parameters while being pulled away from the parameters at a previous training step.  In the supervised setting, retrospective loss results in better test accuracy.

We posit that accelerating critic learning can help actor realise \textit{better} gradient estimates which results in improved policy learning. In lieu with this goal, we propose that using the retrospective loss as a regularizer in the critic objective will accelerate critic learning. In Section~5, we present experiments that demonstrate improved policy learning for the actor achieved with the use of retrospective loss as a regularizer in the critic objective. Furthermore, in Section~6 we observe that even for slowly moving targets (ground truths), as is the case for the critic, retrospective loss yields convergence acceleration.

%To substantiate this, we reinterpret the retrospective loss as a contraction and relate it to the Banach fixed point theorem in Appendix.

\subsection{Formalising SARC}
As the popular Soft Actor-Critic (SAC) algorithm offers both the qualities of sample efficient learning as well as stability \cite{haarnoja2018soft} (Section~\ref{section:related-work}), we therefore, specifically extend SAC in this work. In principle, retrospective regularization can be incorporated with other actor critic methods as well and would be an interesting direction for future exploration. Our algorithm is summarized in Algorithm~\ref{alg:sarc}, where the basic SAC code is reproduced from SpinningUp~\cite{SpinningUp2018} and the additions are highlighted in \textcolor{blue}{blue}. From an algorithmic perspective, SARC is a minimal modification (3 lines) over SAC. 

%We present extensive empirical analysis in Sections~\ref{section:results} and \ref{section:discussion}.  We state some of these results formally and for claims where we do not have a formal proof, we provide intuitive arguments and empirical evidence. 

\begin{algorithm}
    \begin{algorithmic}[1]
    \STATE Input: initial policy parameters $\theta$, Q-function parameters $\phi_1$, $\phi_2$, empty replay buffer $\mathcal{D}$
    \STATE Set target parameters equal to main parameters $\phi_{\text{targ},1} \leftarrow \phi_1$, $\phi_{\text{targ},2} \leftarrow \phi_2$
    \STATE \textcolor{blue}{Previous Q-function parameters set to initial} \newline {\color{blue}$\phi_{prev,1} \gets \phi_1, \phi_{prev,2} \gets \phi_2$}
    
    \REPEAT
        \STATE Observe state $s$, sample $a \sim \pi_\theta(.|s)$
        \STATE Execute $a$ in environment
        \STATE Observe next state $s'$, reward $r$, and done signal $d$ to indicate whether $s'$ is terminal
        \STATE Store $(s, a, s', r, d)$ in replay buffer $D$
        \STATE If $d$ = 1 then $s'$ is terminal, reset environment
        \IF{it's time to update}
            \FOR{$j$ in range numUpdates}
            \STATE Sample batch, $B$ = ${(s, a, s', r, d)}$ 
            from $D$
            \STATE Compute targets for Q-functions
            \begin{align*}
                    & \tilde{a}' \sim \pi_{\theta}(\cdot|s') \\
                    & y (r,s',d) = r + \gamma (1-d)
                    \left(\min_{i=1,2} Q_{\phi_{\text{targ}, i}} (s', \tilde{a}')
                     - \alpha \log \pi_{\theta}(\tilde{a}'|s')\right)
            \end{align*}
            \vspace{-.5cm}
            % \STATE Compute MSE Loss for Q-functions
            \begin{align*}
            % \resizebox{.9\columnwidth}{!}
            &\text{MSE-Loss}(s, a, s', r, d) = 
            \left( Q_{\phi_i}(s,a) - y(r,s',d) \right)^2
            \text{, for } i=1,2
            \end{align*}
            \vspace{-.5cm}
            % \STATE \textcolor{blue}{Compute Retrospective Loss for Q-functions}
            % {\color{blue}\begin{align*}
            % \resizebox{.9\columnwidth}{!}{
            % \text{Retro-Loss}(Q_{\phi_i}(s,a),Q_{\phi_{\text{prev}, i}}(s,a) ,y (r,s',d)) \\ \text{for } i=1,2
            % \end{align*}}
            % \vspace{-.5cm}
            \STATE \textcolor{blue}{Update Q-functions}
            {\color{blue}\begin{align*}
                    & \nabla_{\phi_i} \frac{1}{|B|}\Bigl(\sum_{(s,a,r,s',d) \in B} \text{MSE-Loss}(s,a,r,s',d)\\ &+\mathcal{L}_{\texttt{ret}}(\phi_i, \phi_{prev, i}, y(r, s', d))\Bigr) \text{, for } i=1,2
            \end{align*}}
            \vspace{-.5cm}
            \STATE Update policy
            \begin{align*}
            % \resizebox{.9\columnwidth}{!}{
                    \nabla_{\theta} \frac{1}{|B|}\sum_{s \in B}& \Big(\min_{i=1,2} Q_{\phi_i}(s, \tilde{a}_{\theta}(s)) -\\ 
                    &\alpha \log \pi_{\theta} \left(\left. \tilde{a}_{\theta}(s) \right| s\right) \Big),
                    % }
            \end{align*}
            where $\tilde{a}_{\theta}(s)$ is a sample from $\pi_{\theta}(\cdot|s)$ 
            which is differentiable wrt $\theta$ via the reparametrization trick.
            \STATE Update target Q-functions 
            \begin{align*}
                \phi_{targ,i} \gets \rho\phi_{targ,i} + (1 - \rho)\phi_i && \text{for } i=1,2
            \end{align*}
            \STATE \textcolor{blue}{Update previous Q-functions} {\color{blue}\begin{align*}
                \phi_{prev,i} \gets \phi_i && \text{for } i=1,2
            \end{align*}}
            \vspace{-.5cm}
            \ENDFOR
        \ENDIF
    \UNTIL{convergence}
    \end{algorithmic}
    \caption{Soft Actor Retrospective Critic}
    \label{alg:sarc}
\end{algorithm}

\paragraph{Learning in SARC}
Here we reproduce the basic SAC actor and critic learning algorithms~\citep{haarnoja2018soft,SpinningUp2018} with the retrospective regularizer added to the critic loss, which yields the SARC algorithm. The update equations are as given below. The actor is represented by a function $\pi$, parametrized with $\theta$. Given any state $s'$, the action $\tilde{a}'$ is sampled as:
\begin{equation}\label{eq:sarc-pol}
    \tilde{a}' \sim \pi_{\theta}(\cdot|s').
\end{equation}
The targets for the two critics in SARC are then computed as in SAC:
\begin{align} \label{eq:sarc-critic-target}
    y (r,s', d) = r + \gamma(1-d)
    \Bigl(\min_{i=1,2} &Q_{\phi_{\text{targ}, i}} (s', \tilde{a}')
    - \alpha \log \pi_{\theta}(\tilde{a}'|s')\Bigr)
\end{align}
where $y$ is the target, $r$ is the per-step reward function, $d$ is the done signal, $\phi_{i}$ are the critic parameters, and $\phi_{\text{targ, i}}$ are the corresponding critic parameters for the target. The loss for the critic is then estimated as:
\begin{align} \label{eq:sarc-critic-loss}
\mathcal{L} &= \bigl( Q_{\phi_i}(s,a) - y(r,s', d) \bigr)^2 \notag \\
&+\mathcal{L}_{\texttt{ret}}(\phi_i, \phi_{prev, i}, y(r, s', d))
            \text{, for } i=1,2
\end{align}
where $\mathcal{L}_{\texttt{ret}}$ is given by~\eqref{eq:eq:method-loss-ret}. And finally, the actor is updated as in the original SAC algorithm.

%The critic in actor-critic RL can be aligned with supervised learning, albeit with the presence of moving targets. To motivate the subsequent discussion and our results, we first empirically highlight the generalization of the retrospective regularizer to moving targets. Specifically, we validate this position with results in Section 6 which show that adding the retrospective loss. 

\section{Experimental Setup}\label{section:expt_setup}

To demonstrate that SARC outperforms SAC and is competitive with existing state-of-the-art actor-critic methods such as TD3 and DDPG, we conduct an exhaustive set of experiments. For all of our experiments, we use the implementations provided in SpinningUp \cite{SpinningUp2018}. SpinningUp is an open-source deep RL library.

We modify the original implementation of SAC, provided in SpinningUp, to SARC. Modifying an existing SAC implementation to SARC is minimal and straightforward, requiring addition of a few lines and essentially no compute as presented in Algorithm \ref{alg:sarc}. Within SARC critic loss, for the retrospective regularizer, we use $\kappa = 2$ across all tasks, environments and analyses. 

We conduct our evaluations and analysis across various continuous control tasks. Concretely, we use 6 tasks provided in the DeepMind Control Suite \cite{tassa2018deepmind} and 5 tasks provided in PyBullet Environments \cite{coumans2016pybullet}. For all experiments, we retain all of the default hyperparameters specified originally in the SpinningUp library and let each agent train for 2 million timesteps in the given environment. In our experiments, the same hyperparameters worked for all environments and tasks. The actor and critic network size used in all experiments is [256, 256], unless specified otherwise. This is the default value for the network size specified originally in the SpinningUp library. All models were trained using CPUs on a Linux machine running Ubuntu 16.04 with 2nd Generation Intel Xeon Gold Processor.

In Section~\ref{section:results}, we compare our proposed algorithm SARC to SAC, TD3 and DDPG across 6 tasks provided in the DeepMind Control Suite. In Section~\ref{section:discussion}, we compare retrospective loss with an alternate baseline technique for critic acceleration and analyse our hypothesis of faster critic convergence. We also compare SARC with SAC under various modified experimental settings to further establish our claim. We make all of our code available at: \url{https://github.com/sukritiverma1996/SARC}. We also make all of our data used to construct the graphs and report the figures available at: \url{https://github.com/sukritiverma1996/SARC}.

\section{Results}\label{section:results}

\begin{figure*}
  \centering
    \begin{minipage}[b]{.32\linewidth}
        \centering
        \subcaption{Walker-Stand}
        \includegraphics[width=\linewidth]{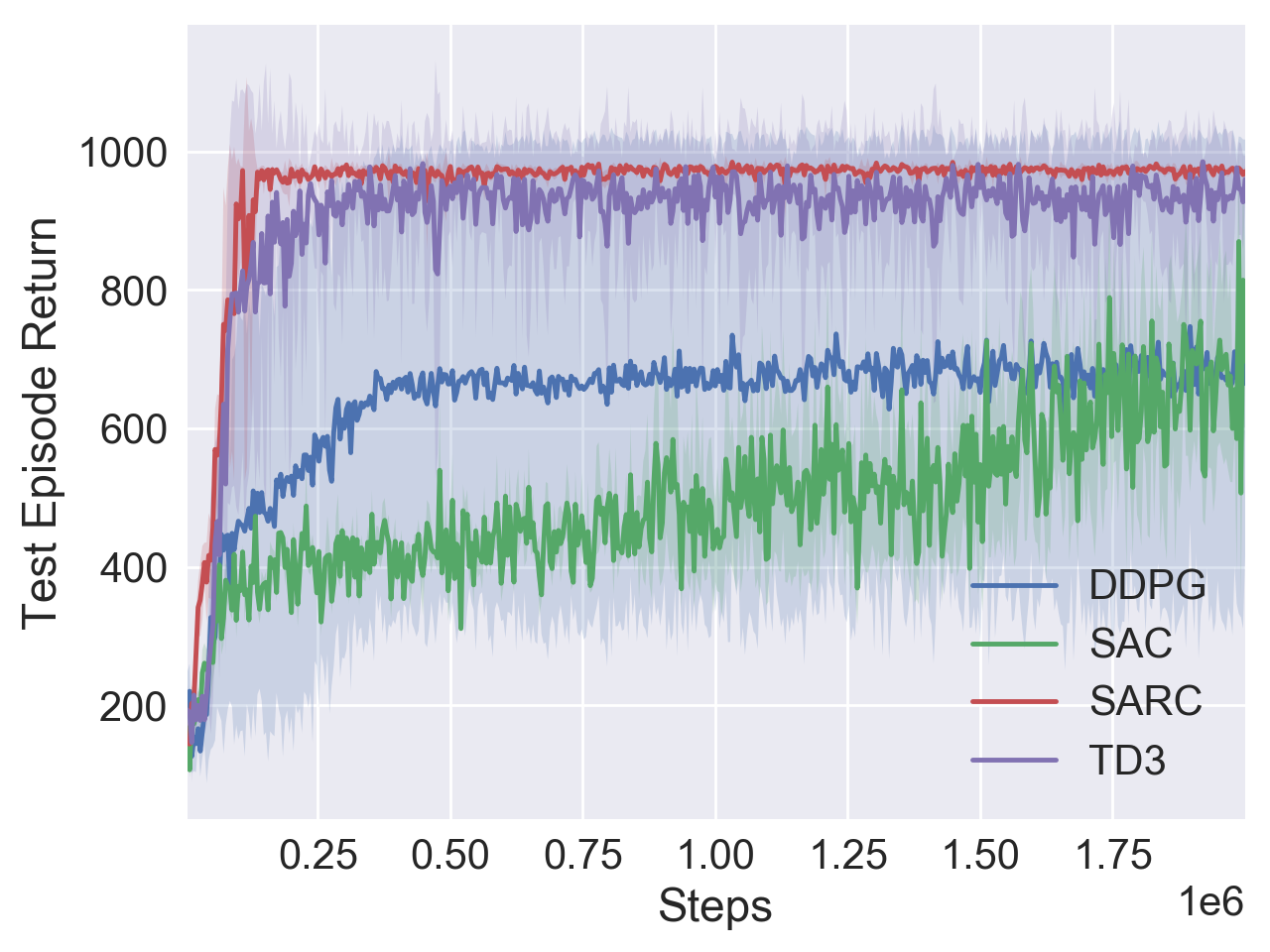}
    \end{minipage}\hfil%
    \begin{minipage}[b]{.32\linewidth}
        \centering
        \subcaption{Walker-Walk}
        \includegraphics[width=\linewidth]{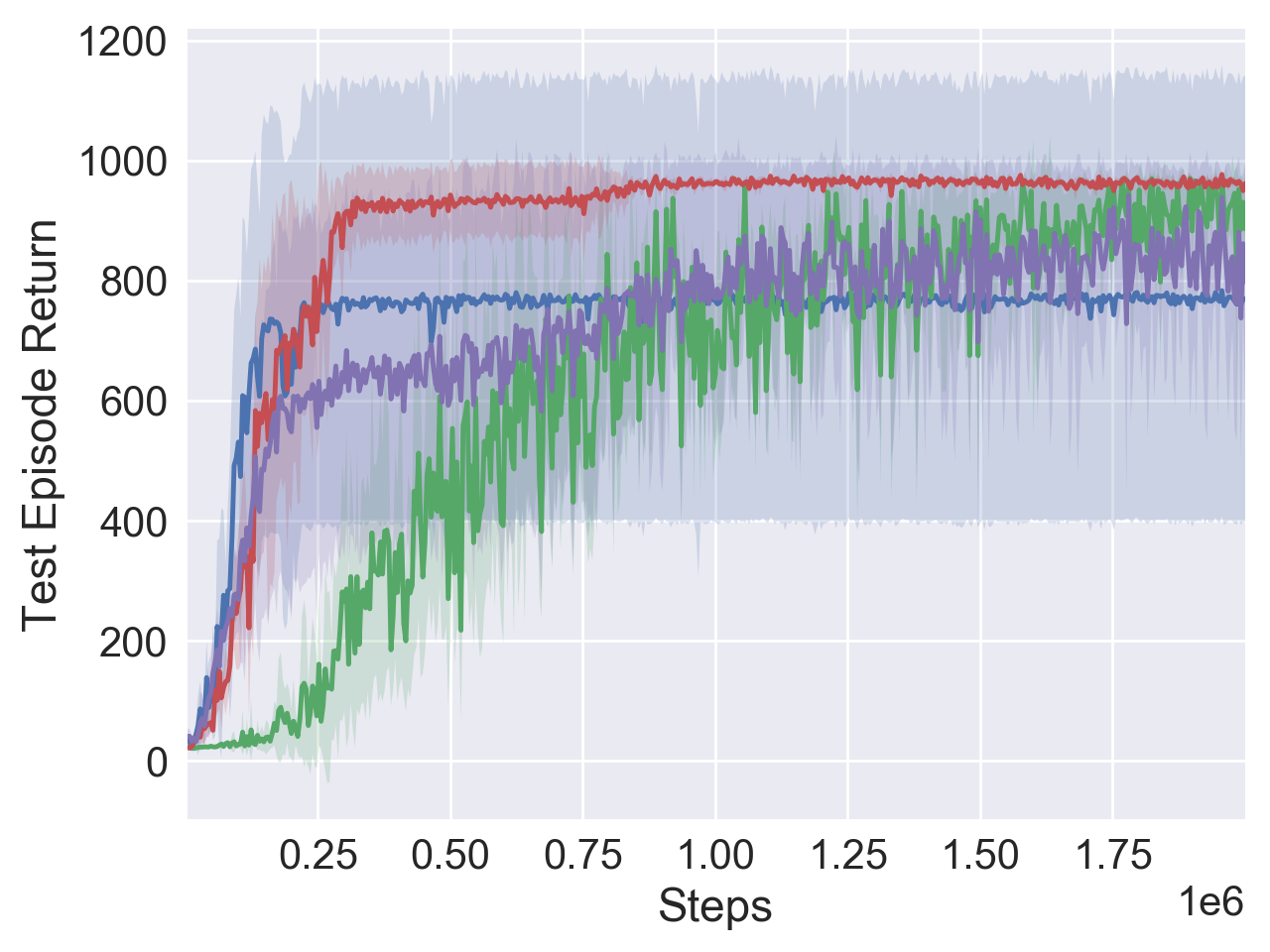}
    \end{minipage}%
    \begin{minipage}[b]{.32\linewidth}
        \centering
        \subcaption{Finger-Spin}
        \includegraphics[width=\linewidth]{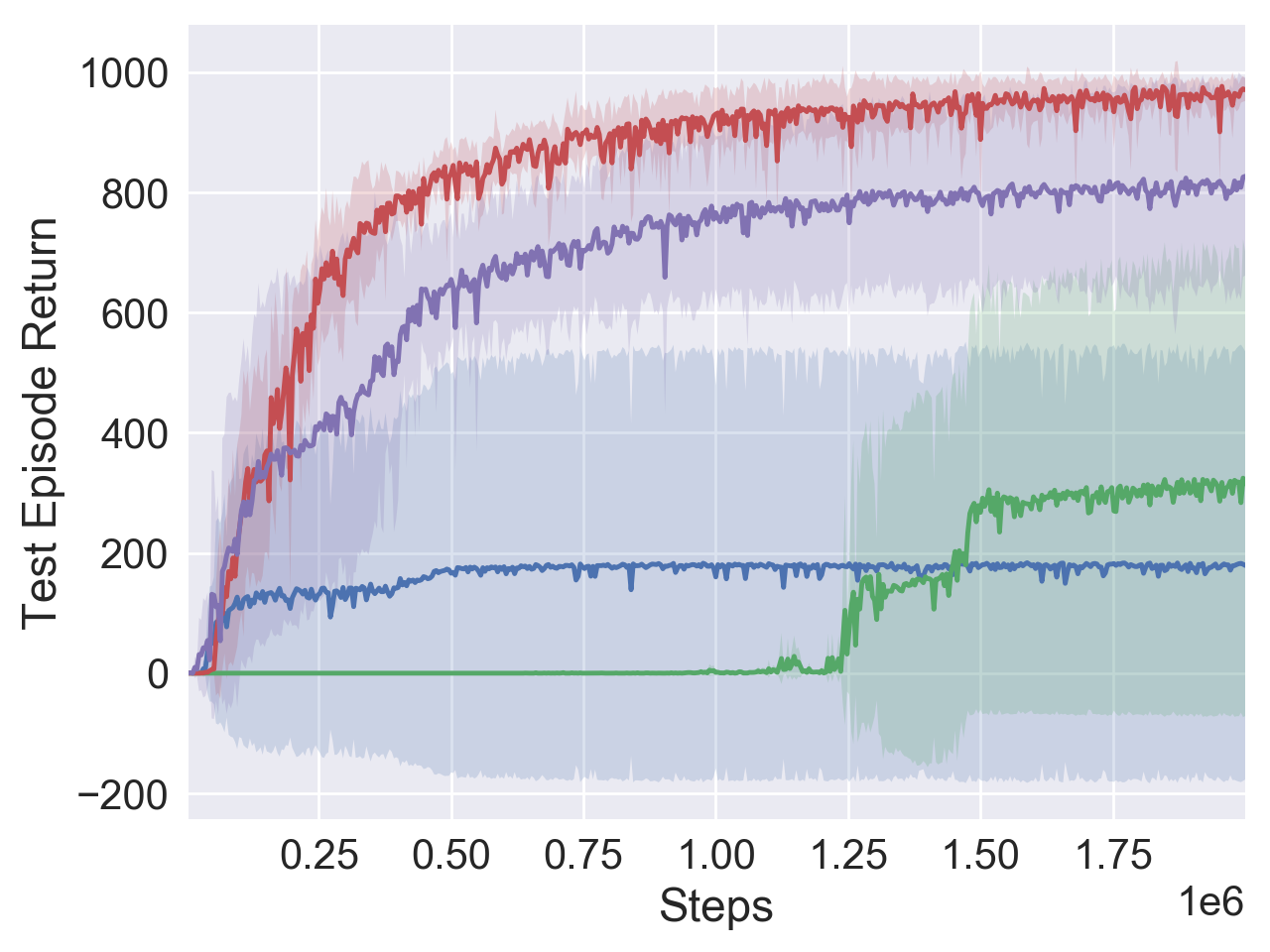}
    \end{minipage}\hfil%
    \begin{minipage}[b]{.32\linewidth}
        \centering
        \subcaption{Cheetah-Run}
        \includegraphics[width=\linewidth]{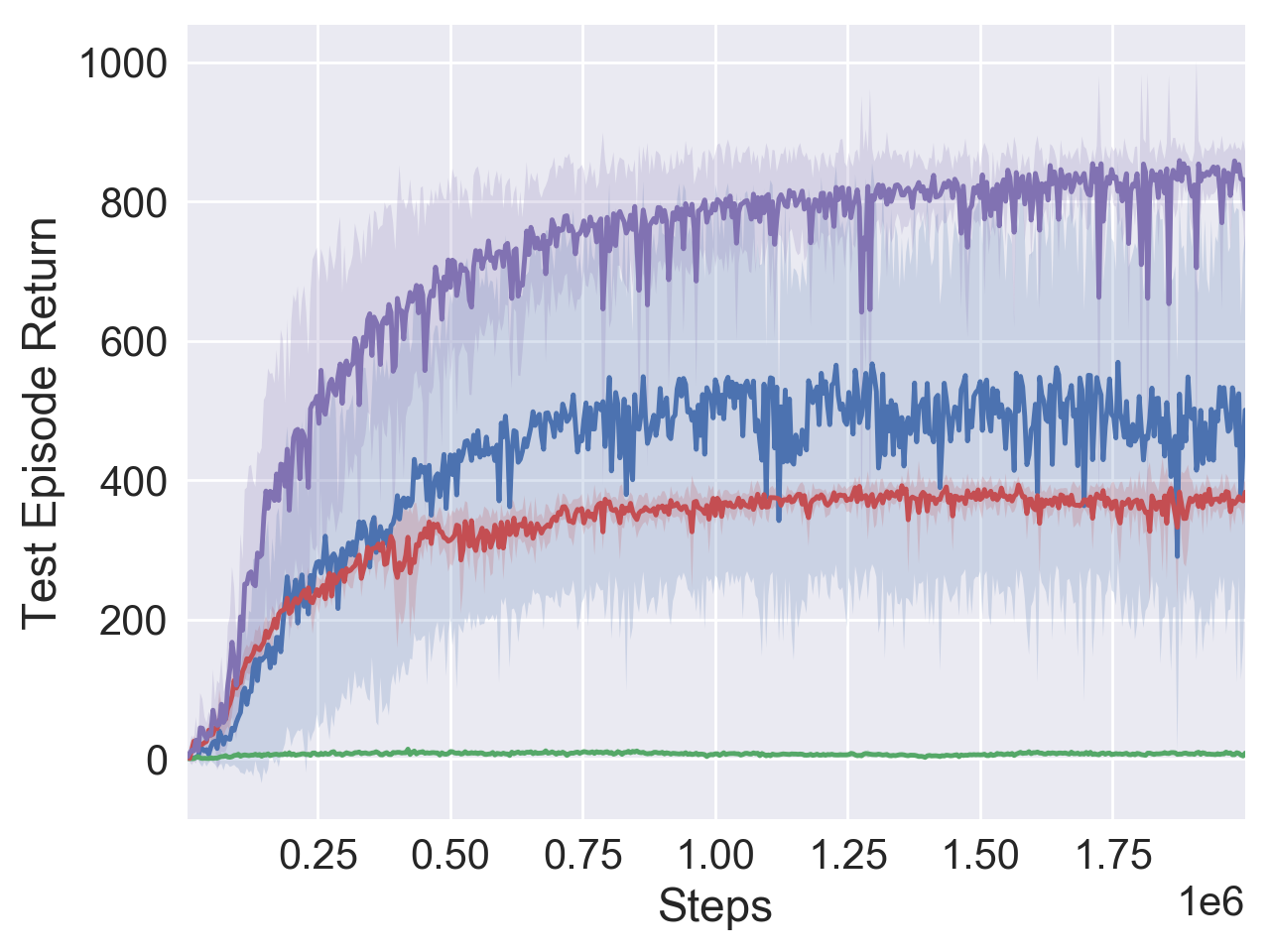}
    \end{minipage}\hfil%
    \begin{minipage}[b]{.32\linewidth}
        \centering
        \subcaption{Reacher-Easy}
        \includegraphics[width=\linewidth]{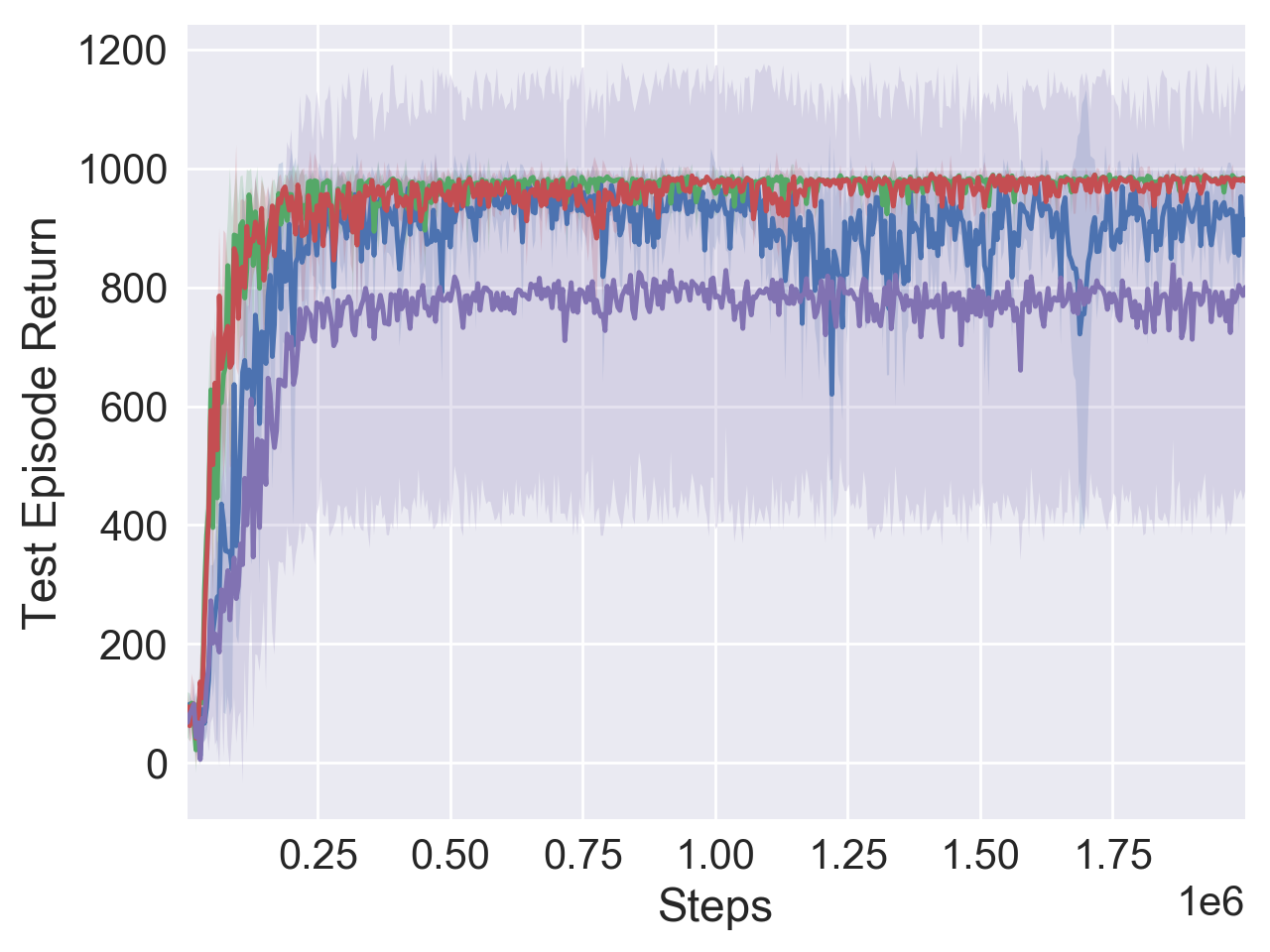}
    \end{minipage}\hfil%
    \begin{minipage}[b]{.32\linewidth}
        \centering
        \subcaption{Reacher-Hard}
        \includegraphics[width=\linewidth]{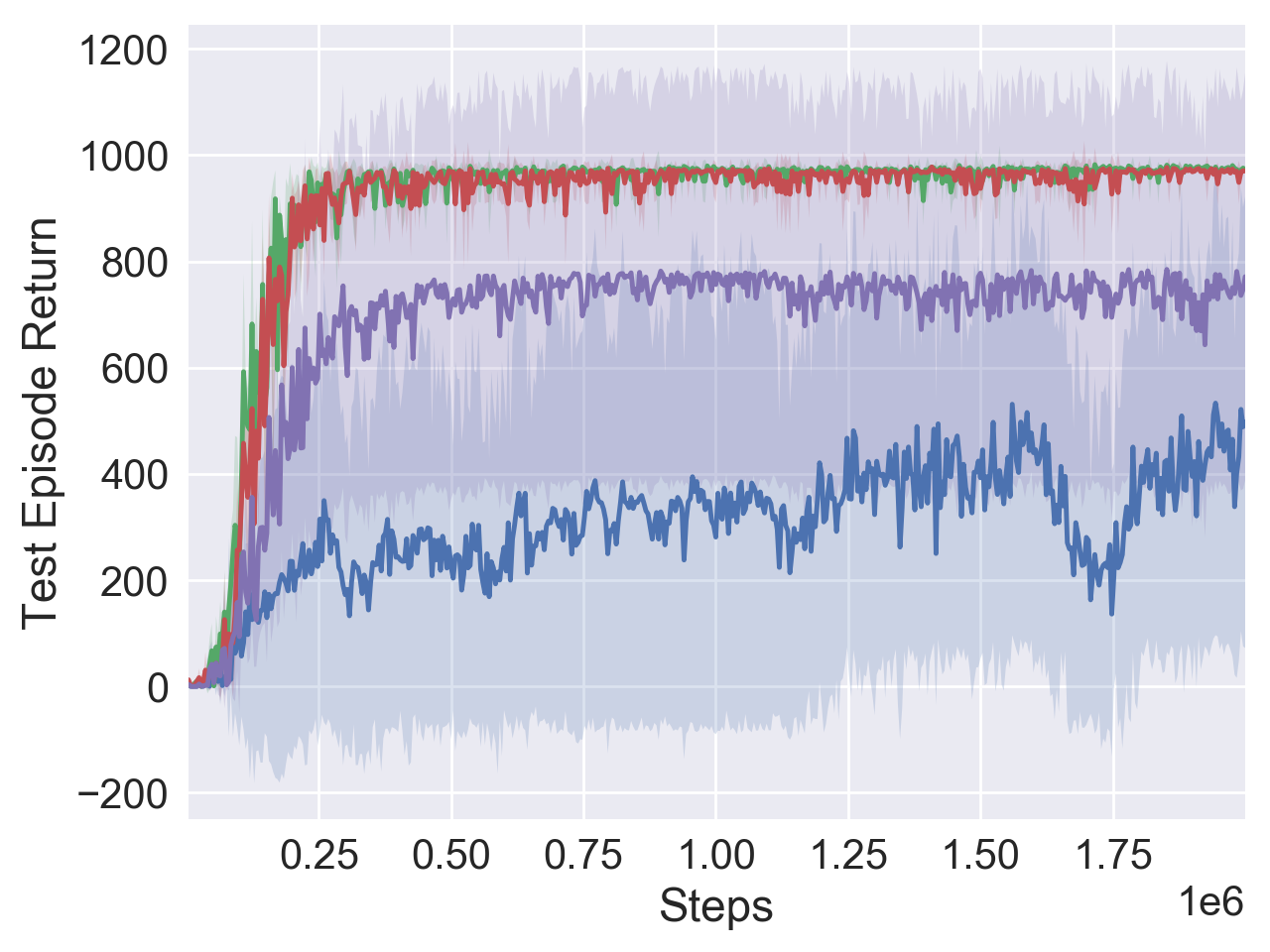}
    \end{minipage}\hfil%
\caption{Results for SARC \textcolor{red}{(red)}, SAC \textcolor{ForestGreen}{(green)}, TD3 \textcolor{Fuchsia}{(purple)} and DDPG \textcolor{blue}{(blue)} across 6 DeepMind Control Suite Tasks. The x-axis shows the timesteps in each environment. The y-axis shows the average value and standard deviation band of return across 5 seeds (10 test episode returns per seed). It can be observed that SARC improves SAC in all environments. SARC also outperforms TD3 and DDPG in most environments.}
\label{fig:spinningup-dmcontrol}
\end{figure*}

In this section, we present comparisons between SARC, SAC, TD3 and DDPG. We perform these evaluations on 6 DeepMind Control Suite tasks: Cheetah-Run, Finger-Spin, Walker-Walk, Walker-Stand, Reacher-Easy, Reacher-Hard. To ensure fairness, consistency and, to rule out outlier behavior, we run five trials for each experiment presented in this paper, each with a different preset random seed. Each experiment uses the same set of 5 seeds.

%The set of these 6 environments together provide contrasting dynamics, varying sizes of observation and action spaces, and varying levels of instability and difficulty. 

\begin{table}
    \caption{Test episode returns of SARC, SAC, TD3 and DDPG on 6 DeepMind Control Suite tasks after training for 2M steps.}
    \label{table:spinningup-dmcontrol}
    \centering
    \footnotesize
    \begin{tabular}{lrrrr}
    \toprule
        \multirow{2}{*}{\textbf{Environment}} & \multicolumn{4}{c}{\textbf{Average Test Episode Return}}\\ \cmidrule(r){2-5}
        
        & \textbf{TD3} & \textbf{DDPG} & \textbf{SAC} & \textbf{SARC}\\
         \midrule

        Walker-Stand & 959.48 & 716.99 & 697.33 & \textbf{970.8}\\
        
        Walker-Walk & 797.87 & 770.08 & 886.28 & \textbf{961.67}\\
        
        Finger-Spin & 827.02 & 179.3 & 315.28 & \textbf{970.94}\\

        Cheetah-Run  & \textbf{789.31} & 500.52 & 8.22 & 382.69 \\
        
        Reacher-Easy & 799.62 & 913.92 & \textbf{978.52} & \textbf{978.32}\\
        
        Reacher-Hard & 771.06 & 497.4 & \textbf{973.08} & \textbf{970.02}\\

        \bottomrule
    \end{tabular}
\end{table}

Table \ref{table:spinningup-dmcontrol} shows the average value across 5 seeds (10 test episode returns per seed) obtained for each of the 6 tasks after training for 2 million timesteps. It can be observed that, at the end of training, SARC outperforms or is at par with existing approaches on \textbf{5 out of 6 environments} presented in \textbf{bold}. 

In figure~\ref{fig:spinningup-dmcontrol}, we present the mean Monte Carlo returns over 10 test episodes at various steps during training. The x-axis shows the timesteps in each environment and the y-axis shows the mean and standard deviation band of above specified returns across 5 seeds.

It can be observed that, on Finger-Spin, Walker-Stand and Walker-Walk, SARC achieves a higher return value faster than any other algorithm. On Reacher-Easy and Reacher-Hard, SARC overlaps with SAC and continues to provide improvement over TD3 and DDPG. SARC substantially improves SAC, particularly on Walker-Stand, Walker-Walk, Cheetah-Run, and Finger-Spin. To ensure and demonstrate that SARC provides a consistent improvement over SAC even under modified experimental settings, we restrict further analyses to these 4 environments. The experiments as described in this section have also been performed on 5 tasks provided in PyBullet Environments \cite{coumans2016pybullet}. These can be found in Appendix A.

%It can be noted that SAC is unable to make any progress and stays at a constant zero return. It can also be noted that on Cheetah-Run, both TD3 and DDPG have a better performance than SARC. It may be the case that the retrospective regulariser within the SARC critic may need some tuning to achieve optimal performance on this particular task.

%SARC is quite similar to SAC, except the modified Critic Loss. Comparing the amount of improvement between SAC and SARC, it may also be noticed that on Cheetah-Run, SAC is unable to perform at all and stays at zero return. The authors think that the inferior performace of SARC may have to do with certain limitations to the amount of Critic acceleration with SARC that is possible given that SAC Critic is unable to train. Such discussions are beyond the scope of this paper. 

\section{Discussion}\label{section:discussion}
First, we explore the effect of alternate baseline strategy to accelerate critic convergence in SAC and compare it with SARC. Next, we present evidence of accelerated critic convergence in SARC. Finally, we demonstrate that SARC significantly outperforms SAC for increased network sizes. We note that, for ease of exposition, we restrict the comparative scope of this analysis to SAC, which is the most similar baseline and also learns a soft critic.

\subsection{Increasing critic update frequency for SAC}
\begin{figure}[b]
  \centering
    \begin{minipage}[b]{.3\linewidth}
        \centering
        \subcaption{Finger-Spin}
        \includegraphics[width=\linewidth]{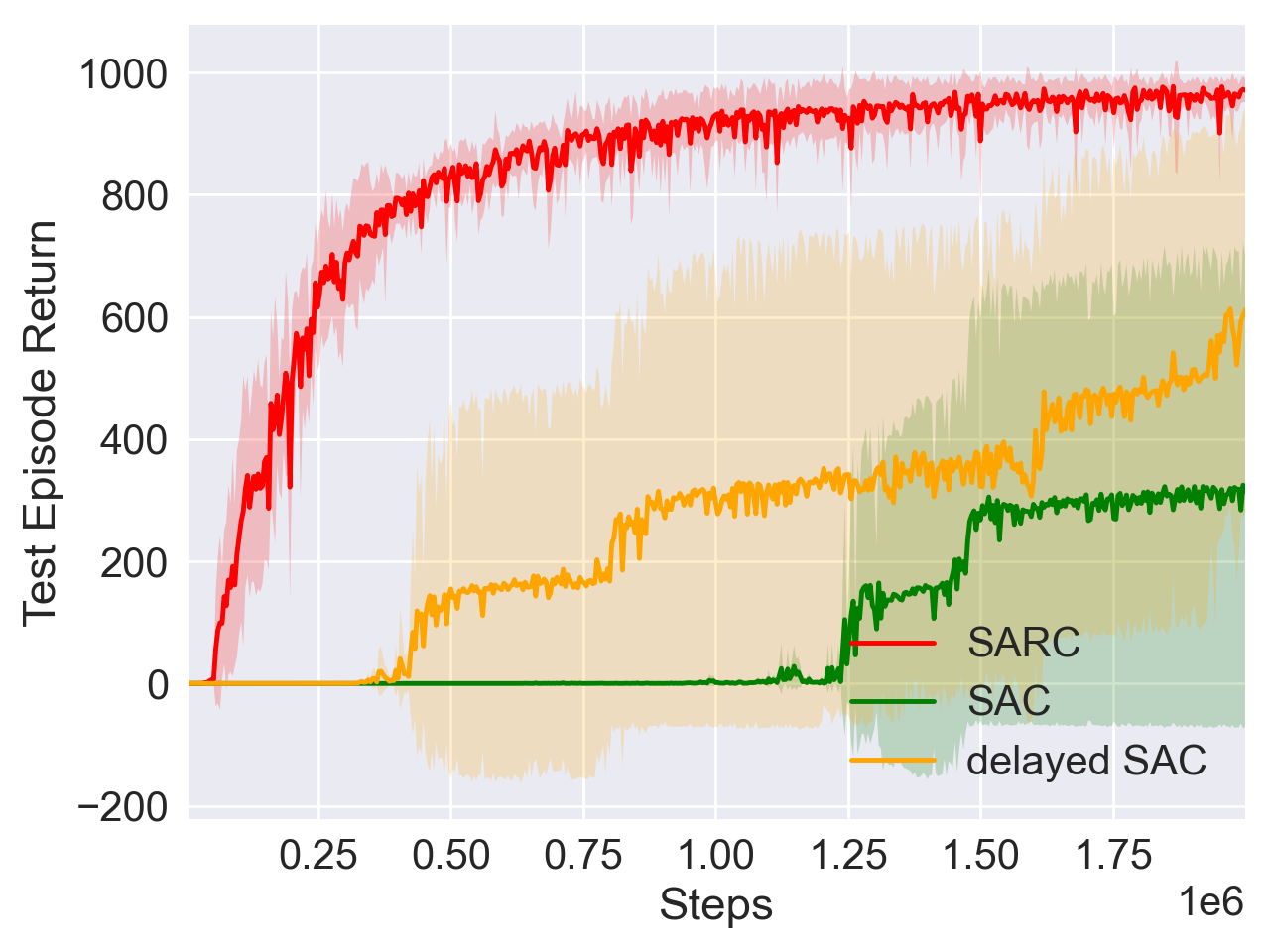}
    \end{minipage}\hfil%
    \begin{minipage}[b]{.3\linewidth}
        \centering
        \subcaption{Cheetah-Run}
        \includegraphics[width=\linewidth]{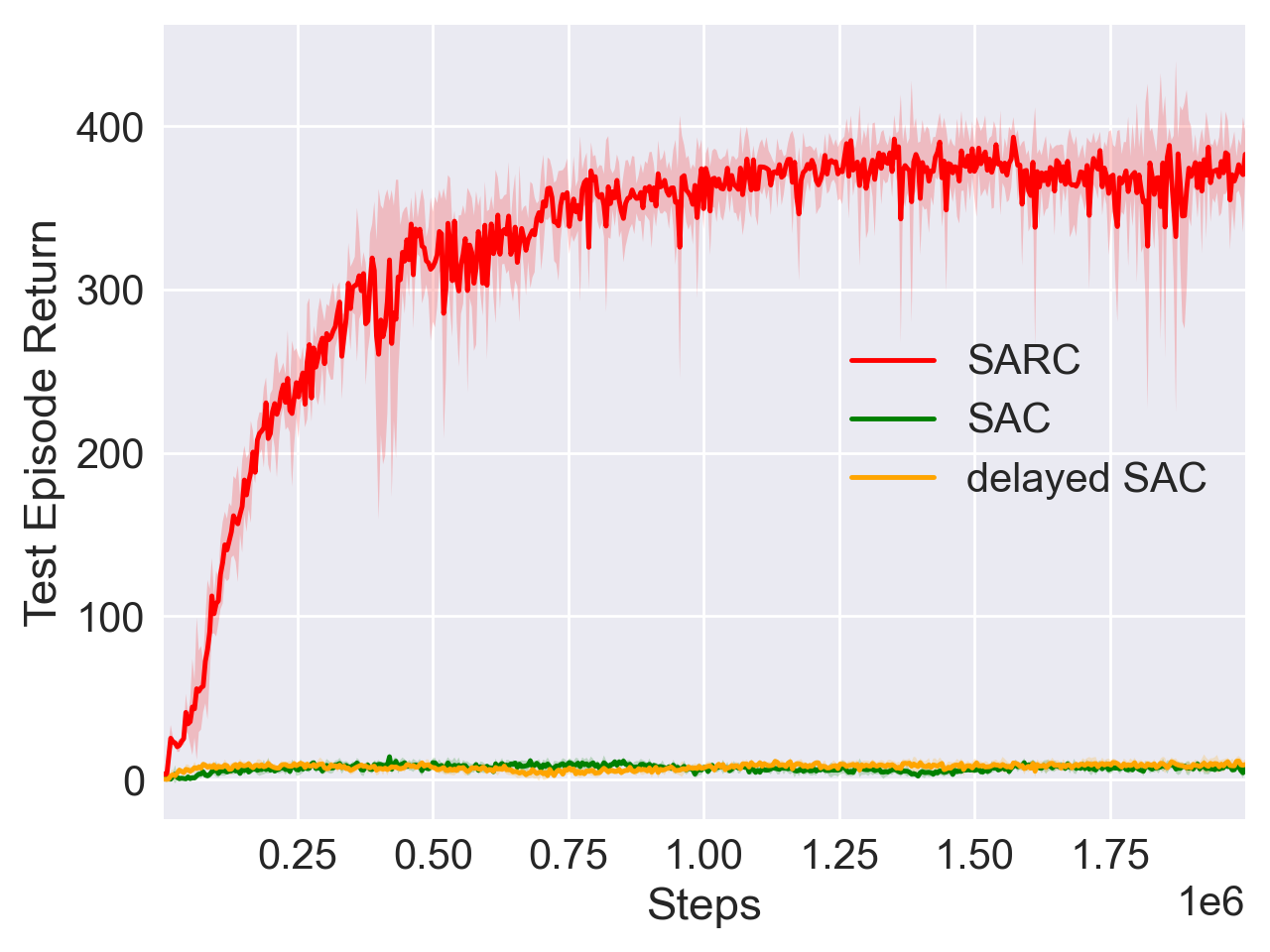}
    \end{minipage}\hfil%
    \begin{minipage}[b]{.3\linewidth}
        \centering
        \subcaption{Walker-Stand}
        \includegraphics[width=\linewidth]{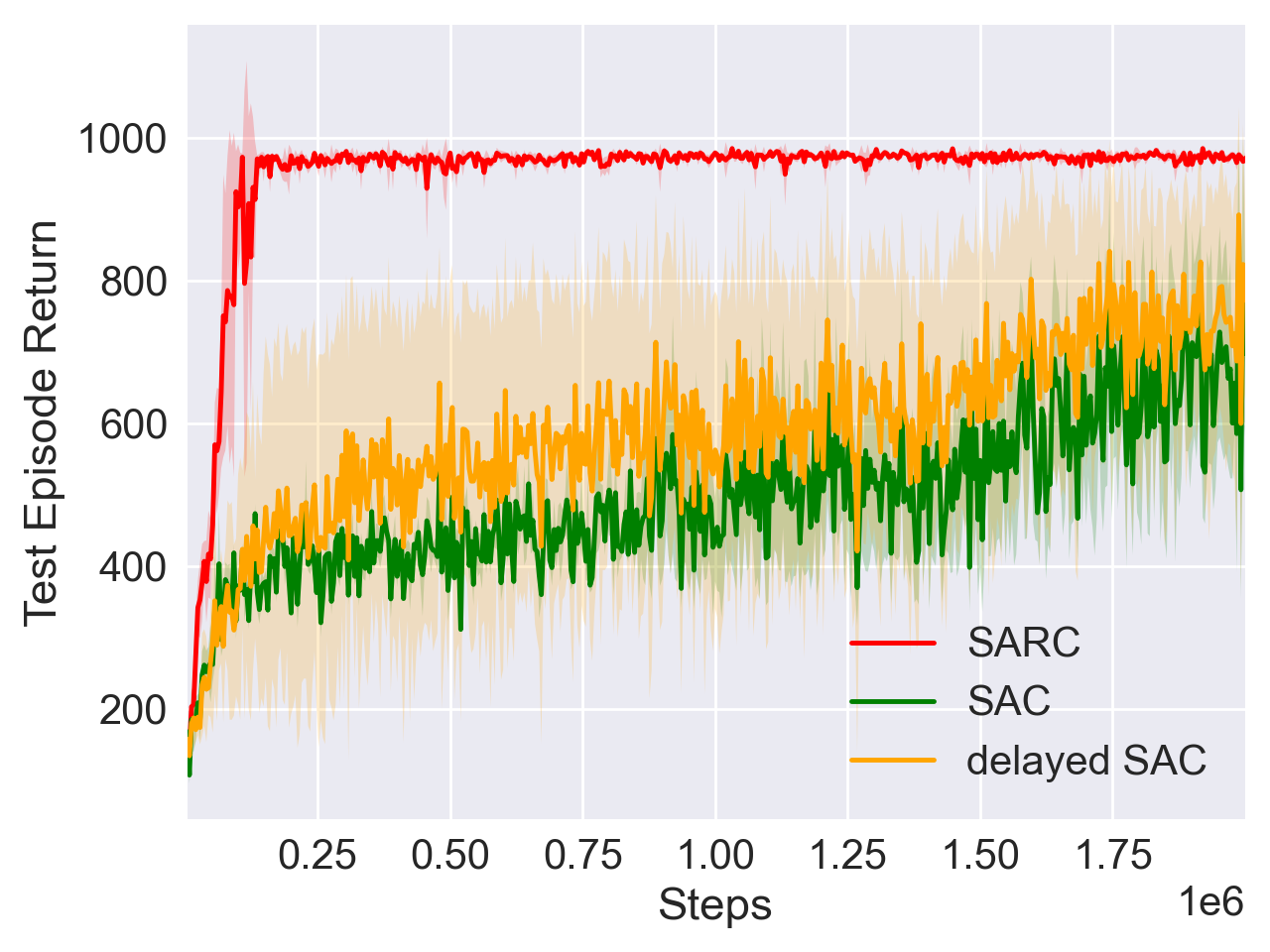}
    \end{minipage}\hfil%
\caption{Comparing SARC and delayed SAC. The x-axis shows the timesteps in each environment. The y-axis shows the average value and standard deviation band of return across 5 seeds (10 test episode returns per seed). Compared to delayed SAC \textcolor{Mycolor1}{(yellow)}, SARC \textcolor{red}{(red)} provides a more reliable and potent improvement over SAC.}
\label{fig:modified-sac}
\end{figure}

An alternate baseline strategy to accelerate convergence of the critic \textit{relative} to the actor can be to update critic more frequently. A similar idea has been proposed in TD3~\cite{fujimoto2018addressing}, where the critic updates more frequently than the actor. 

In this section, we compare SARC against a version of SAC where the critic updates twice as frequently as the actor. We term this as "delayed SAC". We do this for three environments: Finger-Spin, Cheetah-Run and Walker-Stand. Figure \ref{fig:modified-sac} has the results of this investigation. The x-axis shows the timesteps in each environment and the y-axis shows the mean and standard deviation band of Mean Monte Carlo returns over 10 test episodes across 5 seeds. It can be observed that while delayed SAC provides some improvement over original SAC, it is not as reliable and not as potent as the improvement provided by SARC.
%SARC continues to provide consistent improvement over delayed SAC.

\subsection{Modifying entropy regularization coefficient}
In this section, we explore the effect of modifying the entropy regularization coefficient, $\alpha$ in SAC and SARC. 
It may be the case that the significant performance gain that SARC is offering over SAC may disappear if the entropy regularizer, $\alpha$, is tuned.

\begin{figure}
   \centering
    \begin{minipage}[b]{.3\linewidth}
        \centering
        \subcaption{Finger-Spin, $\alpha = 0.1$}
        \includegraphics[width=\linewidth]{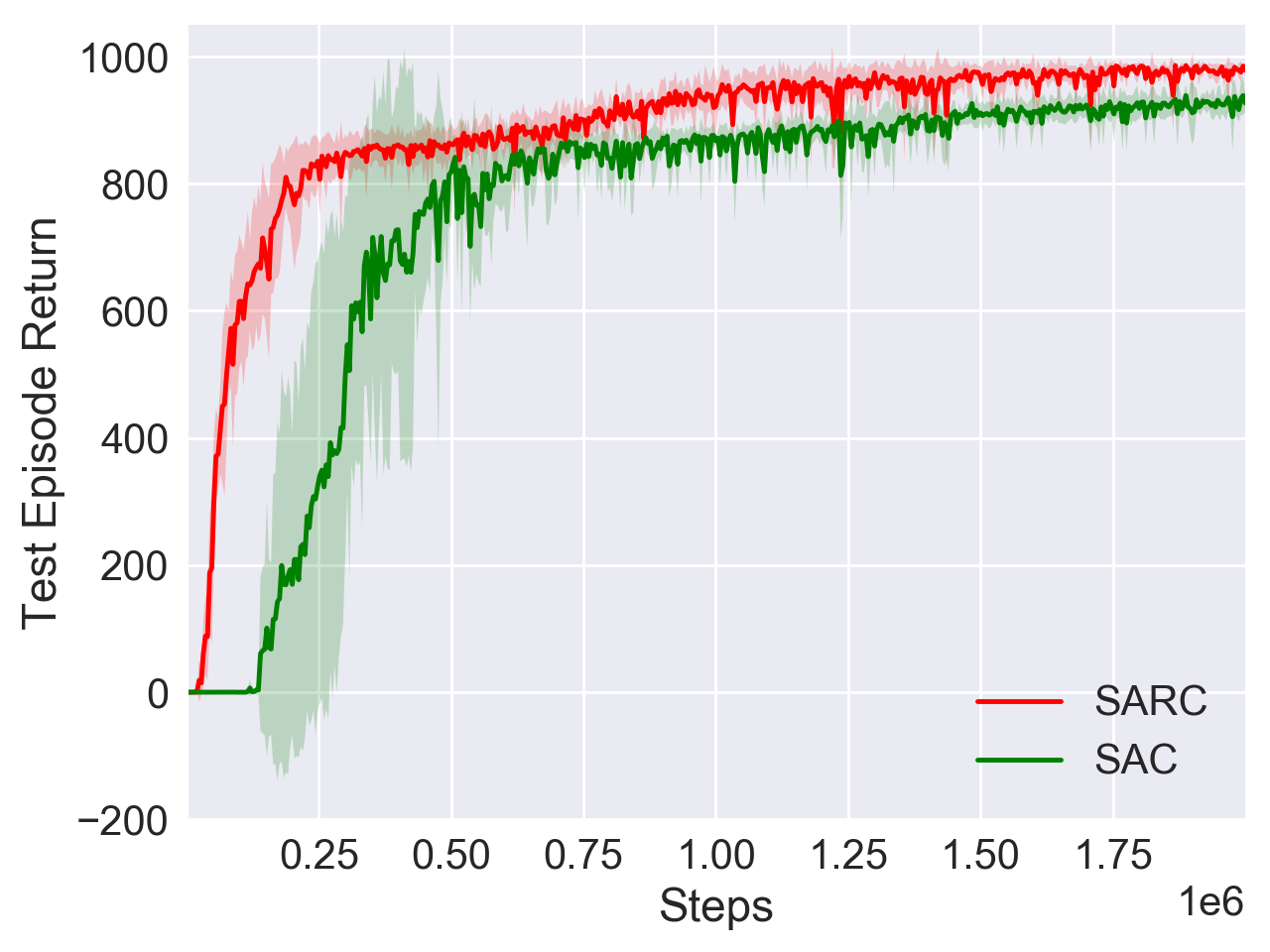}
    \end{minipage}\hfil%
    \begin{minipage}[b]{.3\linewidth}
        \centering
        \subcaption{Walker-Stand, $\alpha = 0.1$}
        \includegraphics[width=\linewidth]{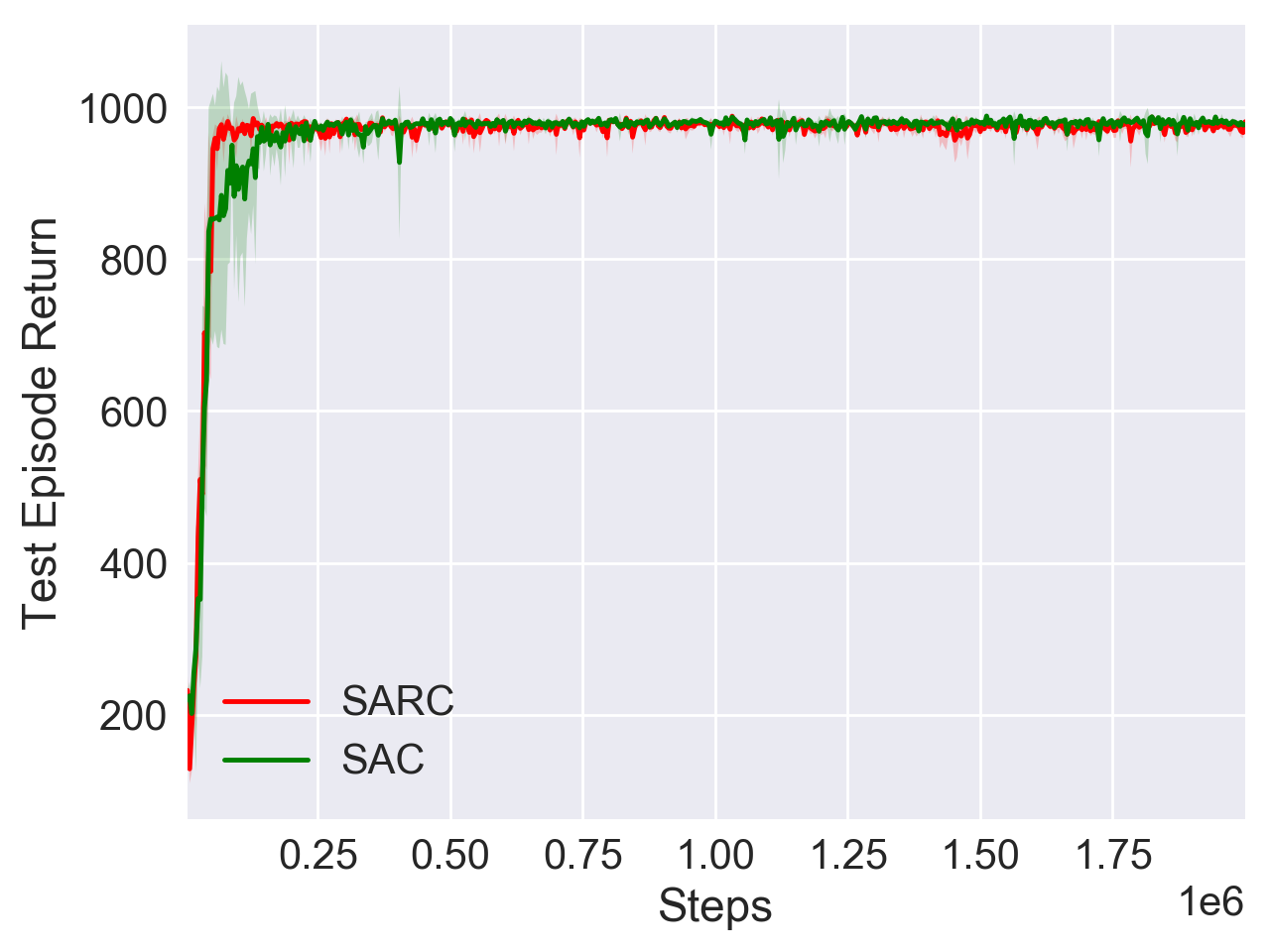}
    \end{minipage}\hfil%
    \begin{minipage}[b]{.3\linewidth}
        \centering
        \subcaption{Walker-Walk, $\alpha = 0.1$}
        \includegraphics[width=\linewidth]{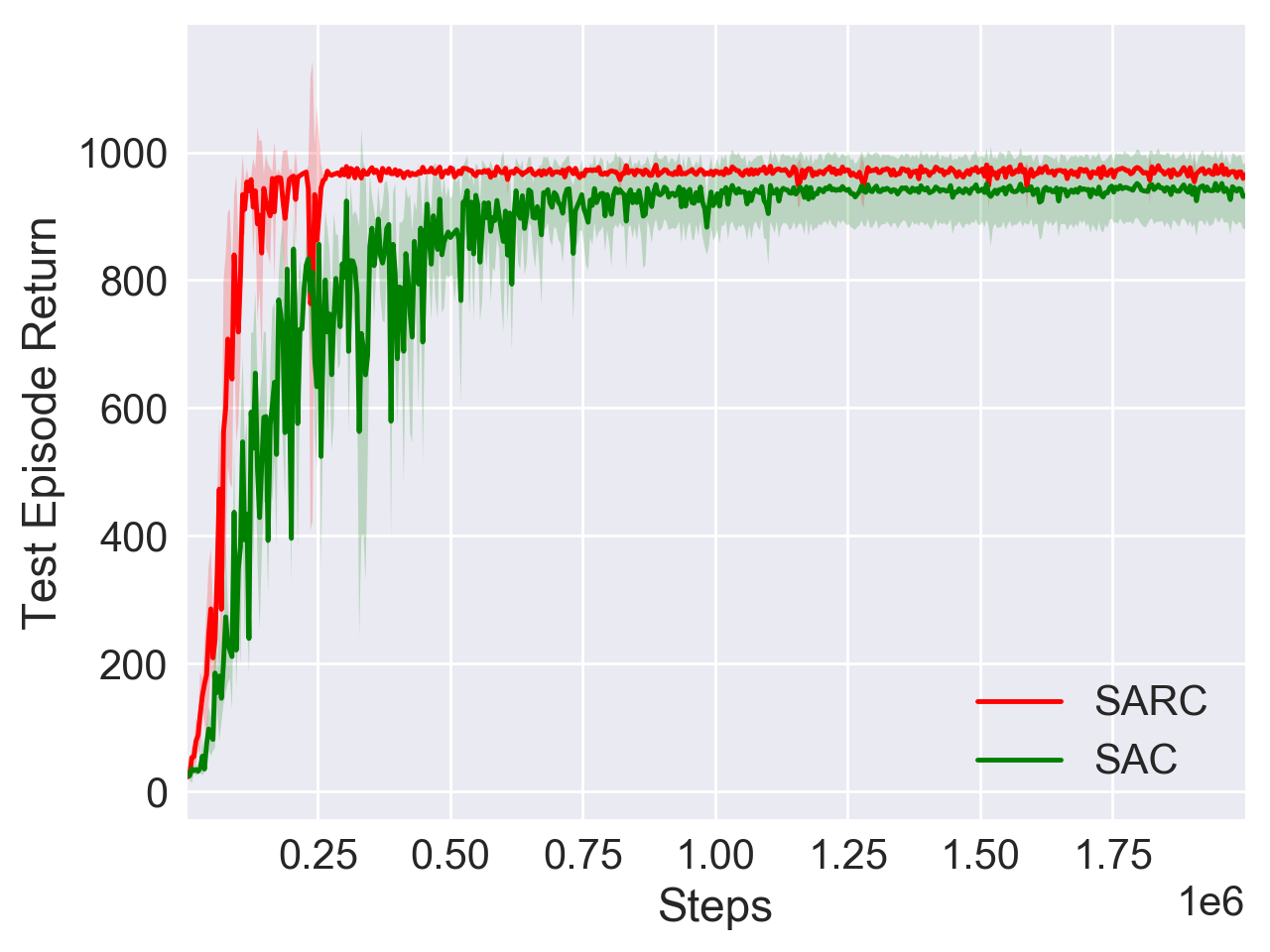}
    \end{minipage}
    \begin{minipage}[b]{.3\linewidth}
        \centering
        \subcaption{Finger-Spin, $\alpha = 0.4$}
        \includegraphics[width=\linewidth]{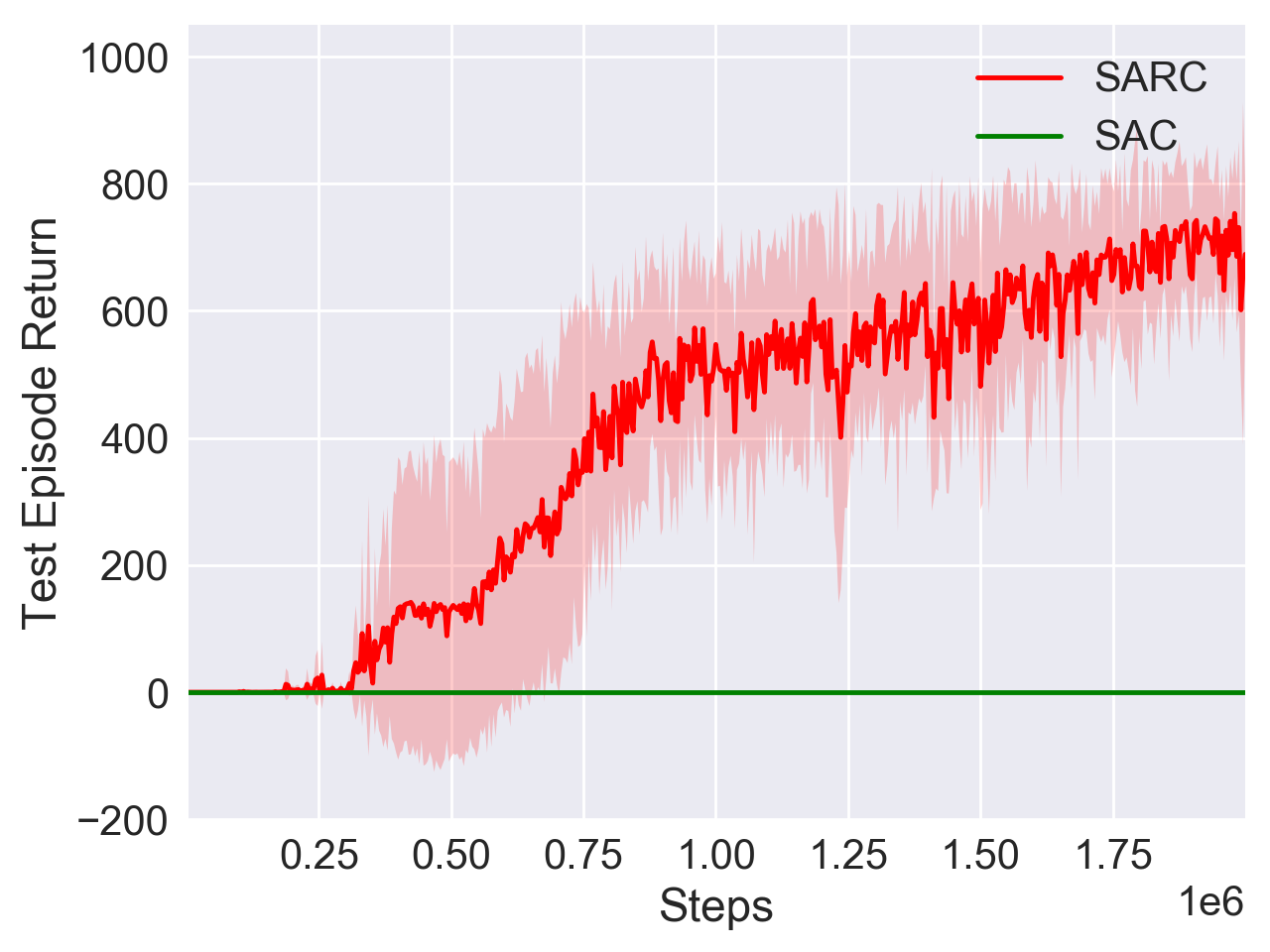}
    \end{minipage}\hfil%
    \begin{minipage}[b]{.3\linewidth}
        \centering
        \subcaption{Walker-Stand, $\alpha = 0.4$}
        \includegraphics[width=\linewidth]{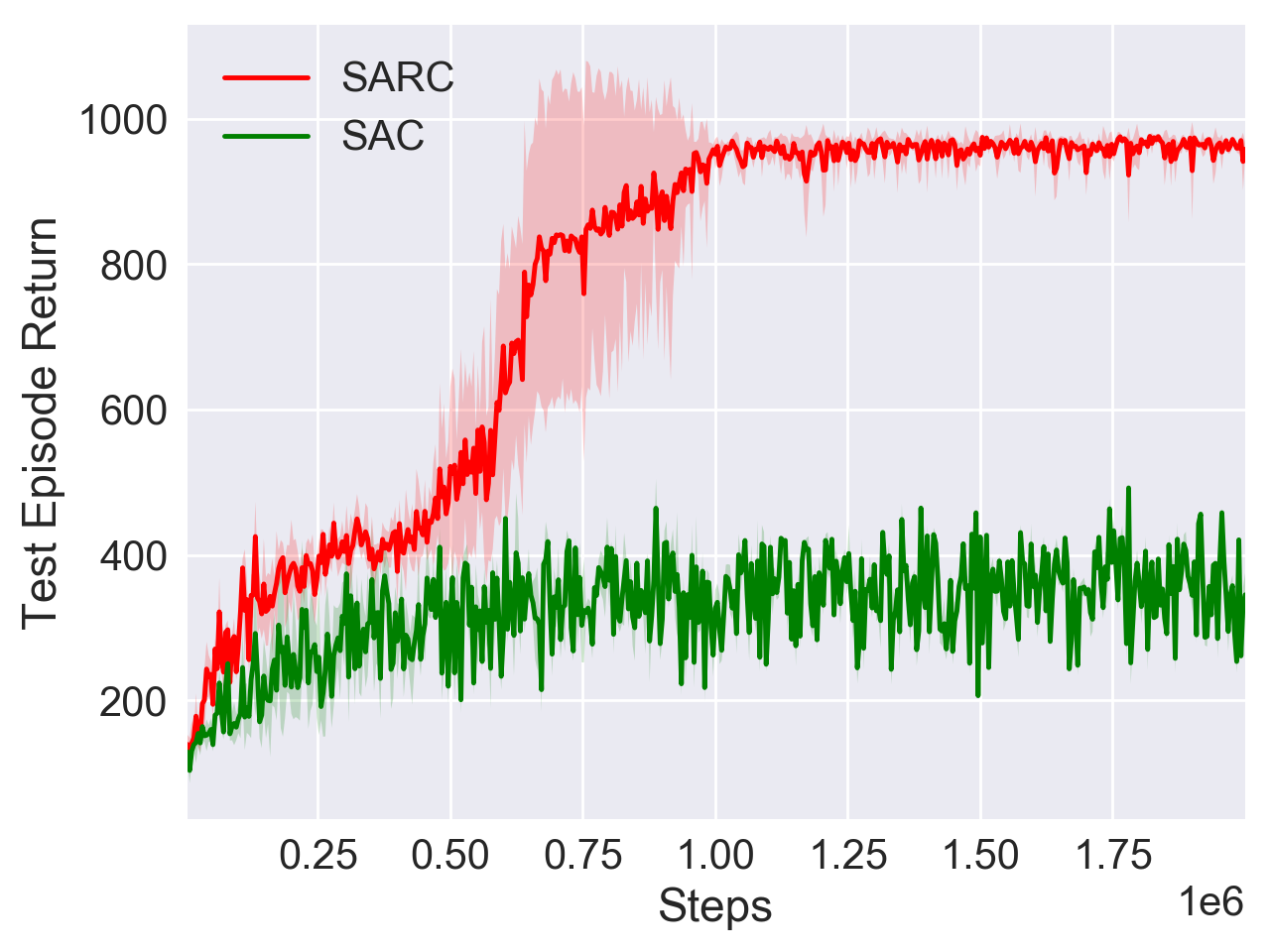}
    \end{minipage}\hfil%
    \begin{minipage}[b]{.3\linewidth}
        \centering
        \subcaption{Walker-Walk, $\alpha = 0.4$}
        \includegraphics[width=\linewidth]{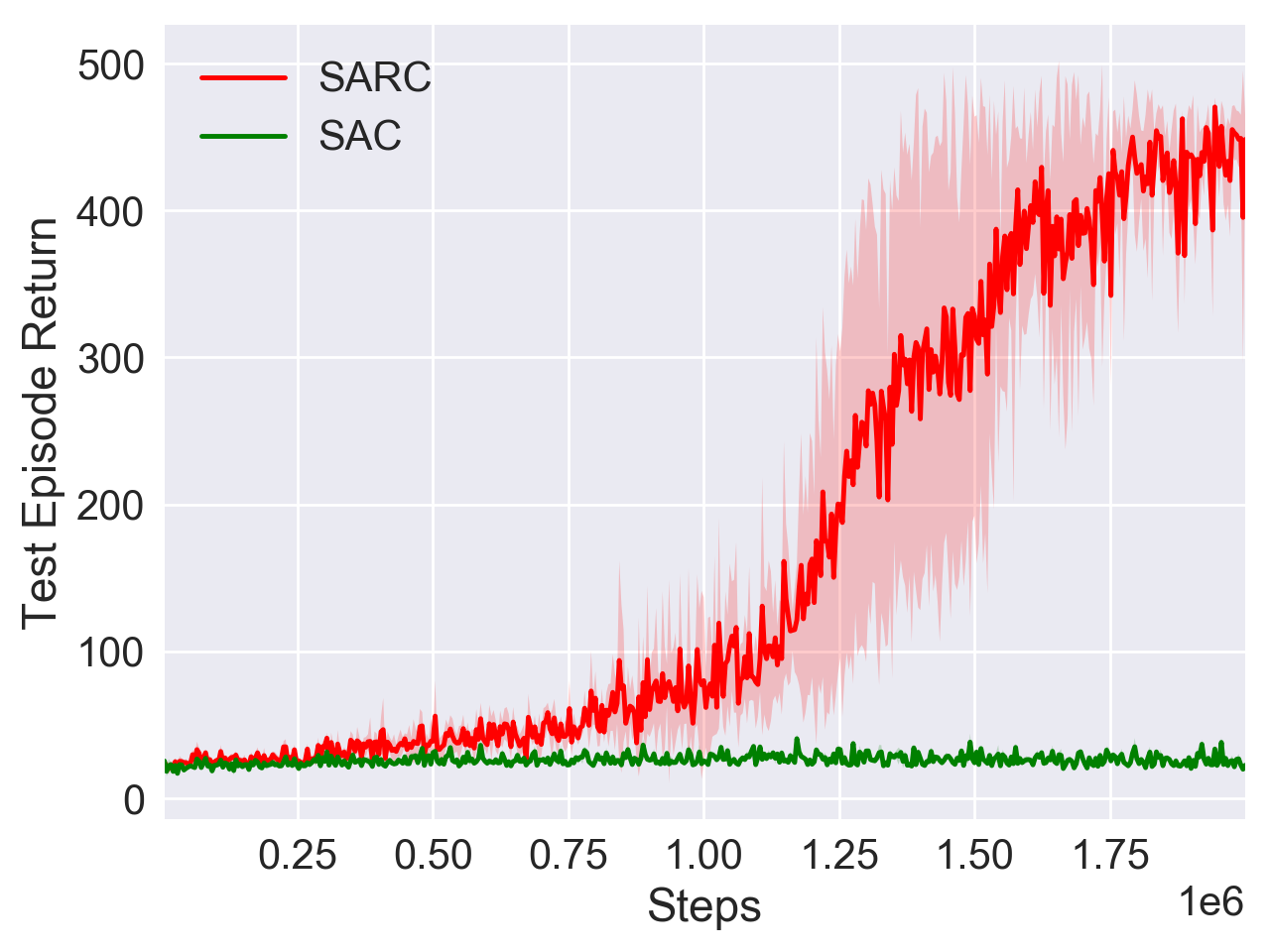}
    \end{minipage}\hfil%
    
\caption{Results for SARC \textcolor {red}{(red)} and SAC on tuning the entropy regularization coefficient, $\alpha$. The x-axis shows the timesteps in each environment. The y-axis shows the average value and standard deviation band of return across 5 seeds (10 test episode returns per seed). It can be seen that SARC outperforms SAC across different values of $\alpha$.}
\label{fig:alpha-tune}
\end{figure}

All experiments presented in Section~\ref{section:results} had value of $\alpha = 0.2$. In this section, we look at two more values of  $\alpha$: 0.1 and 0.4. All other hyperparameters are held at fixed values as described in Section~\ref{section:expt_setup}. We do this for three environments: Finger-Spin, Walker-Stand and Walker-Walk. Figure \ref{fig:alpha-tune} has the results of this investigation. Mean Monte Carlo returns over 10 test episodes at various steps during training have been plotted. The x-axis shows the timesteps in each environment and the y-axis shows the mean and standard deviation band of above specified returns across 5 seeds. It can be observed that SARC continues to provide consistent improvement over SAC across values of $\alpha$.

\subsection{Tracking SARC critic convergence with respect to Monte Carlo policy evaluation}

We train one SAC agent and one SARC agent for 2 million timesteps. During training, we save the intermediate actor and critic at every 10,000 timesteps. Given an intermediate actor-critic pair, we compute the mean squared difference between Monte Carlo estimate of the return of the current policy and Q($s_0$,$a_0$) learned by the critic for that policy. This difference is termed as ``Q-Error''. If the critic value function has converged to the correct value for a given policy, then this difference should be zero. The idea is to track and compare how fast the critic converges.

\begin{figure}
  \centering
    \begin{minipage}[b]{.3\linewidth}
        \centering
        \subcaption{Finger-Spin}
        \includegraphics[width=\linewidth]{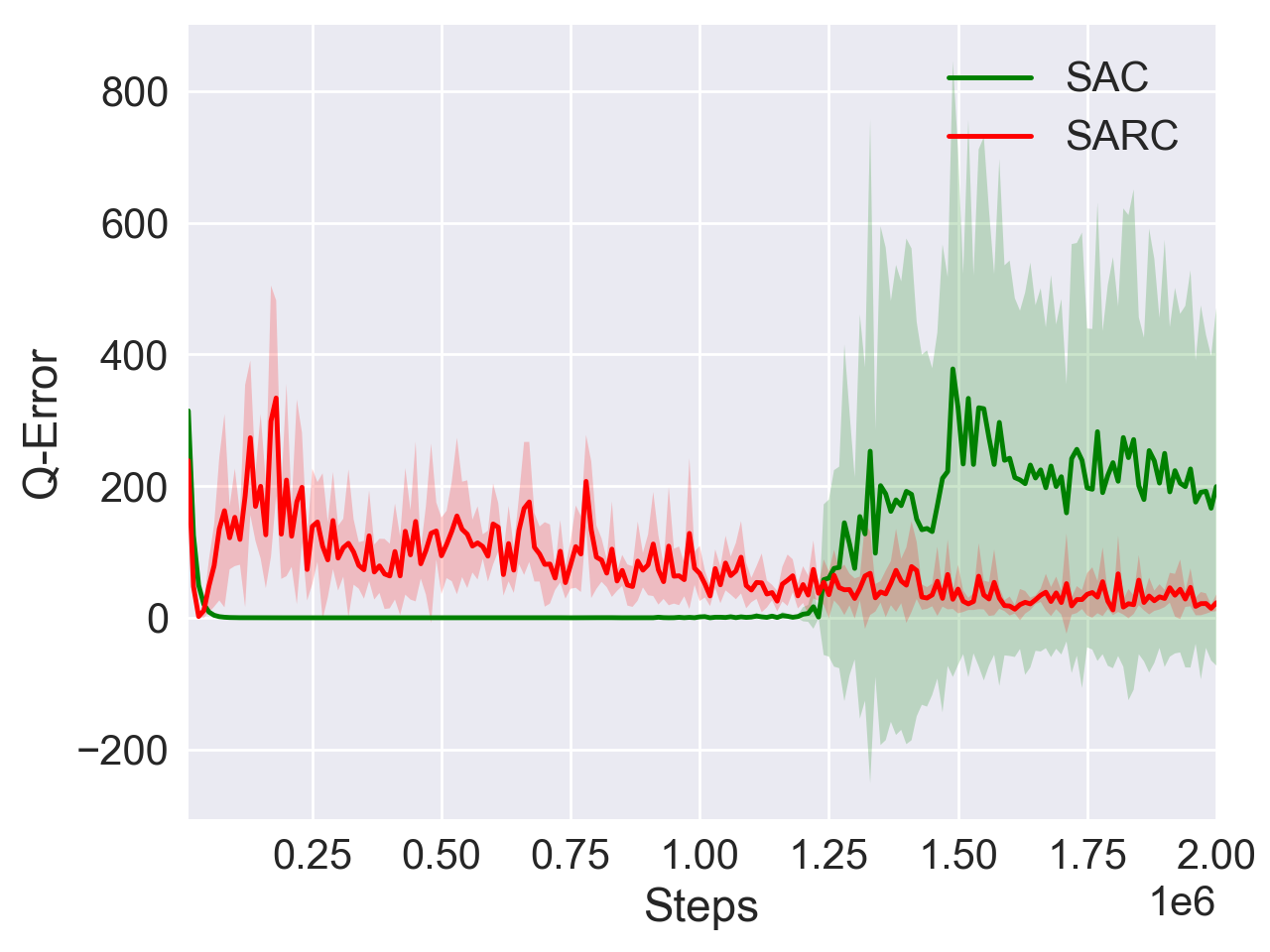}
    \end{minipage}\hfil%
    \begin{minipage}[b]{.3\linewidth}
        \centering
        \subcaption{Cheetah-Run}
        \includegraphics[width=\linewidth]{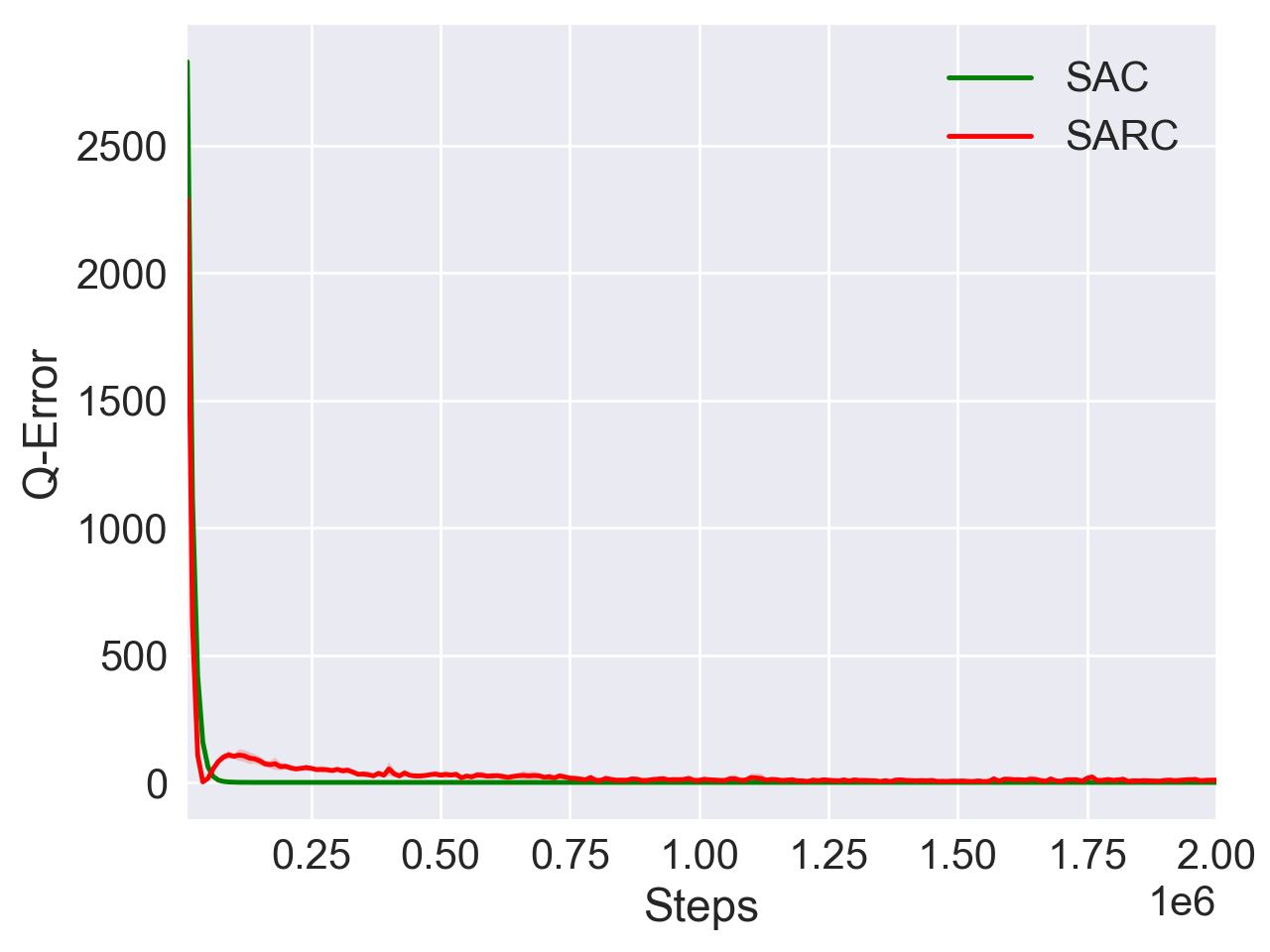}
    \end{minipage}\hfil%
    \begin{minipage}[b]{.3\linewidth}
        \centering
        \subcaption{Walker-Stand}
        \includegraphics[width=\linewidth]{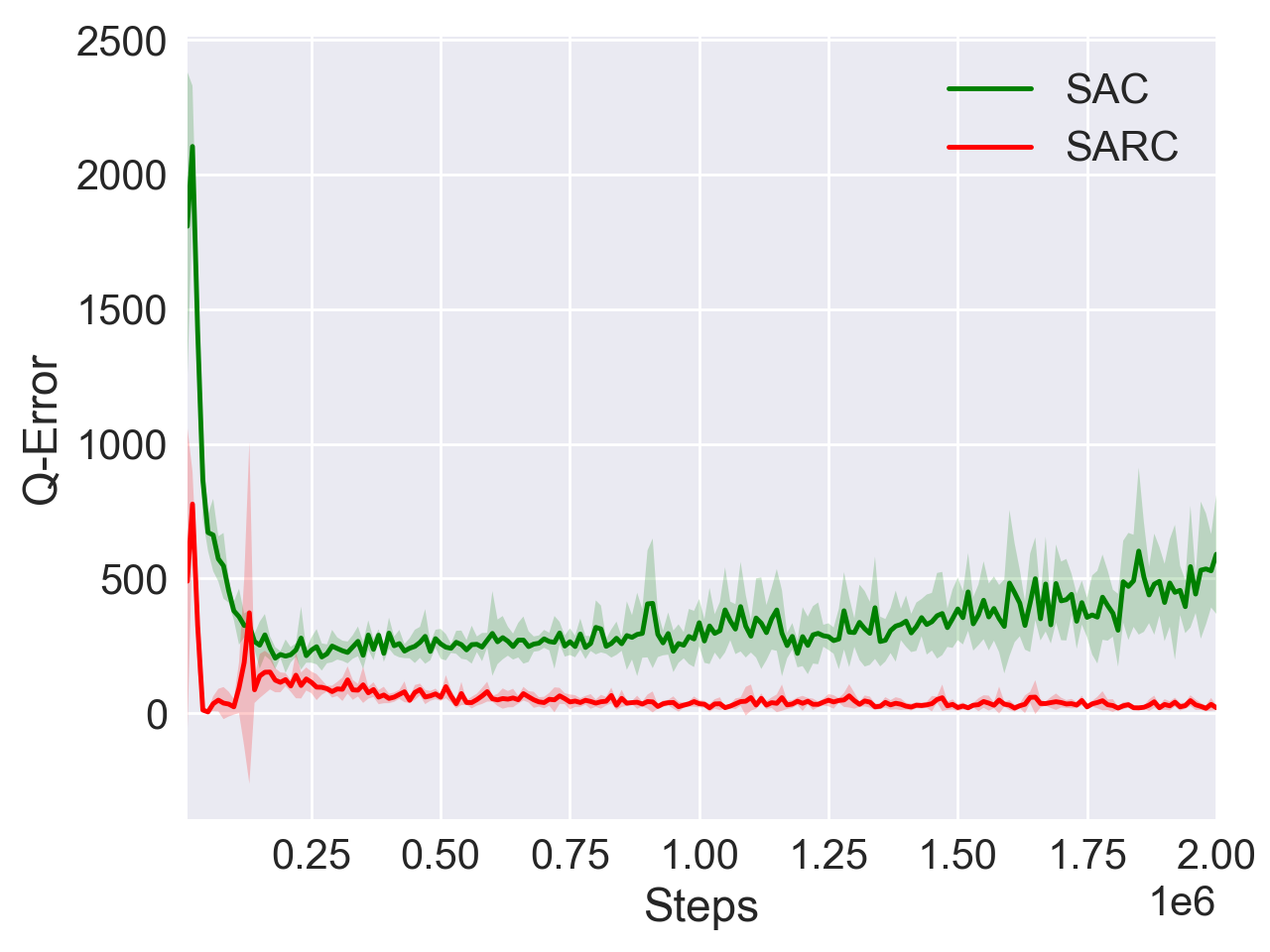}
    \end{minipage}\hfil%
\caption{Tracking critic convergence with respect to Monte Carlo policy evaluation across training. The x-axis shows the timesteps in each environment. The y-axis shows the mean and standard deviation band of ``Q-Error'' across 5 seeds. It can be seen that SARC \textcolor{red}{(red)} critic converges faster.}
\label{fig:critic-convergence}
\end{figure}

We do this for three environments: Finger-Spin, Cheetah-Run, Walker-Stand. Figure \ref{fig:critic-convergence} has the results of this investigation. The average ``Q-Error'' over 5 seeds has been plotted. The graphs also show the standard deviation band across the 5 random seeds. It can be seen that SARC critic converges faster than the SAC critic on Finger-Spin and Walker-Stand.

\subsection{Increasing actor and critic network complexity}

It may be possible that SAC can perform as well as SARC with a bigger network. In this section, we compare SAC and SARC on tasks from DeepMind Control Suite after increasing the network sizes for actor and critic to [400, 300]. This is another commonly found architecture in literature relating to continuous control \cite{henderson2018deep}. In our previous experiments, we were using network size [256, 256]. 

\begin{figure}
   \centering
    \begin{minipage}[b]{.3\linewidth}
        \centering
        \subcaption{Walker-Stand}
        \includegraphics[width=\linewidth]{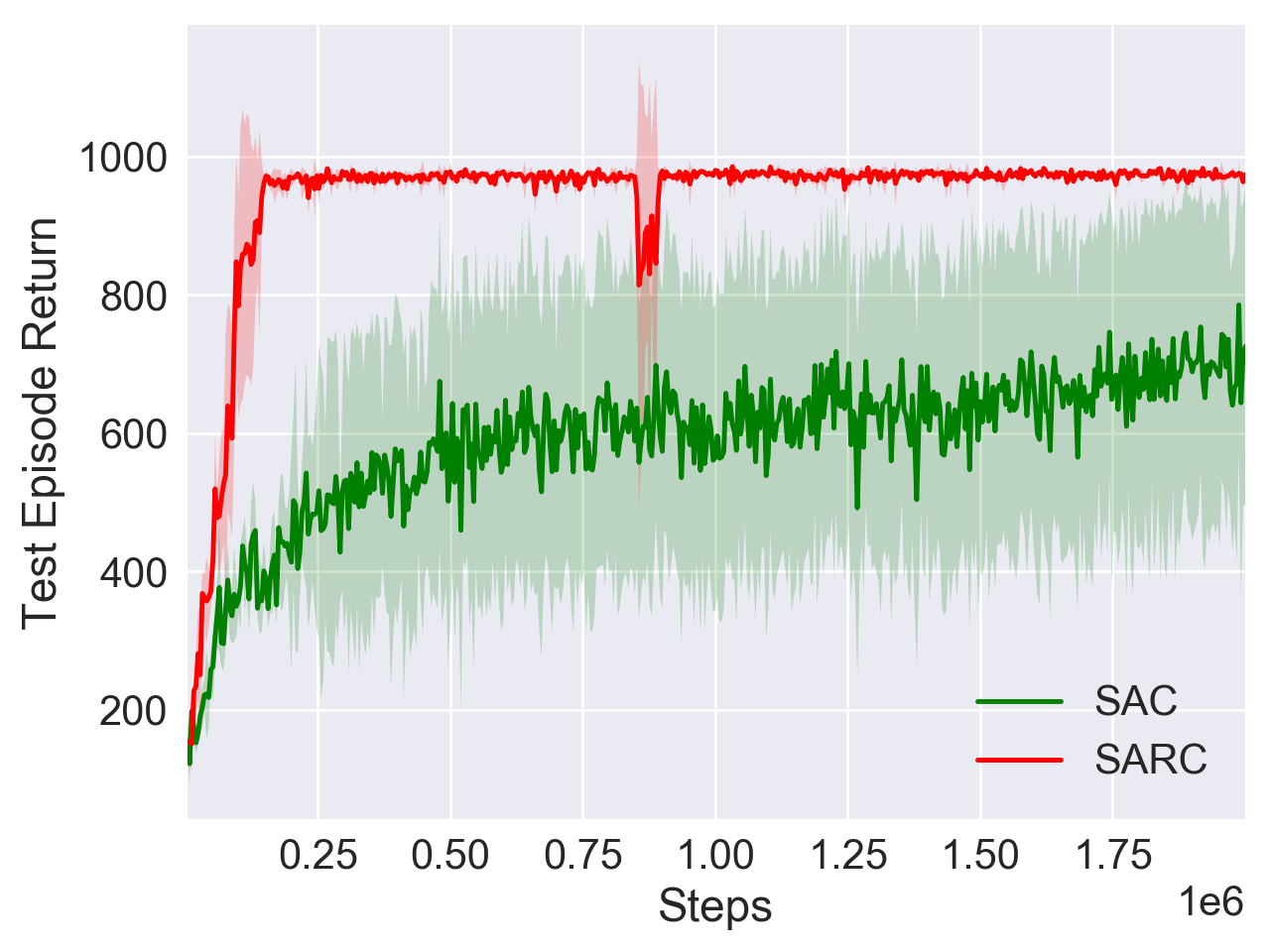}
    \end{minipage}\hfil%
    \begin{minipage}[b]{.3\linewidth}
        \centering
        \subcaption{Walker-Walk}
        \includegraphics[width=\linewidth]{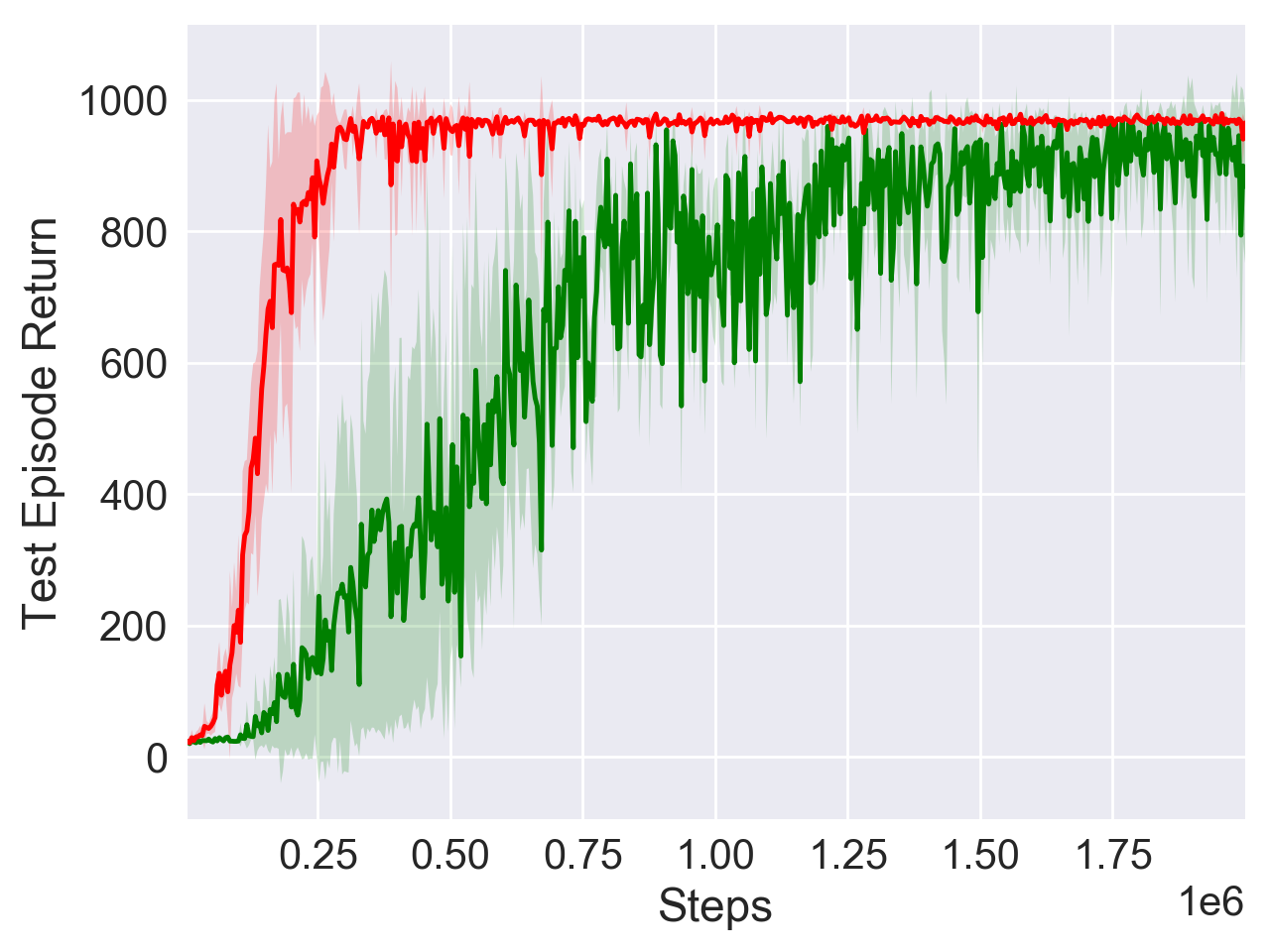}
    \end{minipage}\hfil%
    \begin{minipage}[b]{.3\linewidth}
        \centering
        \subcaption{Finger-Spin}
        \includegraphics[width=\linewidth]{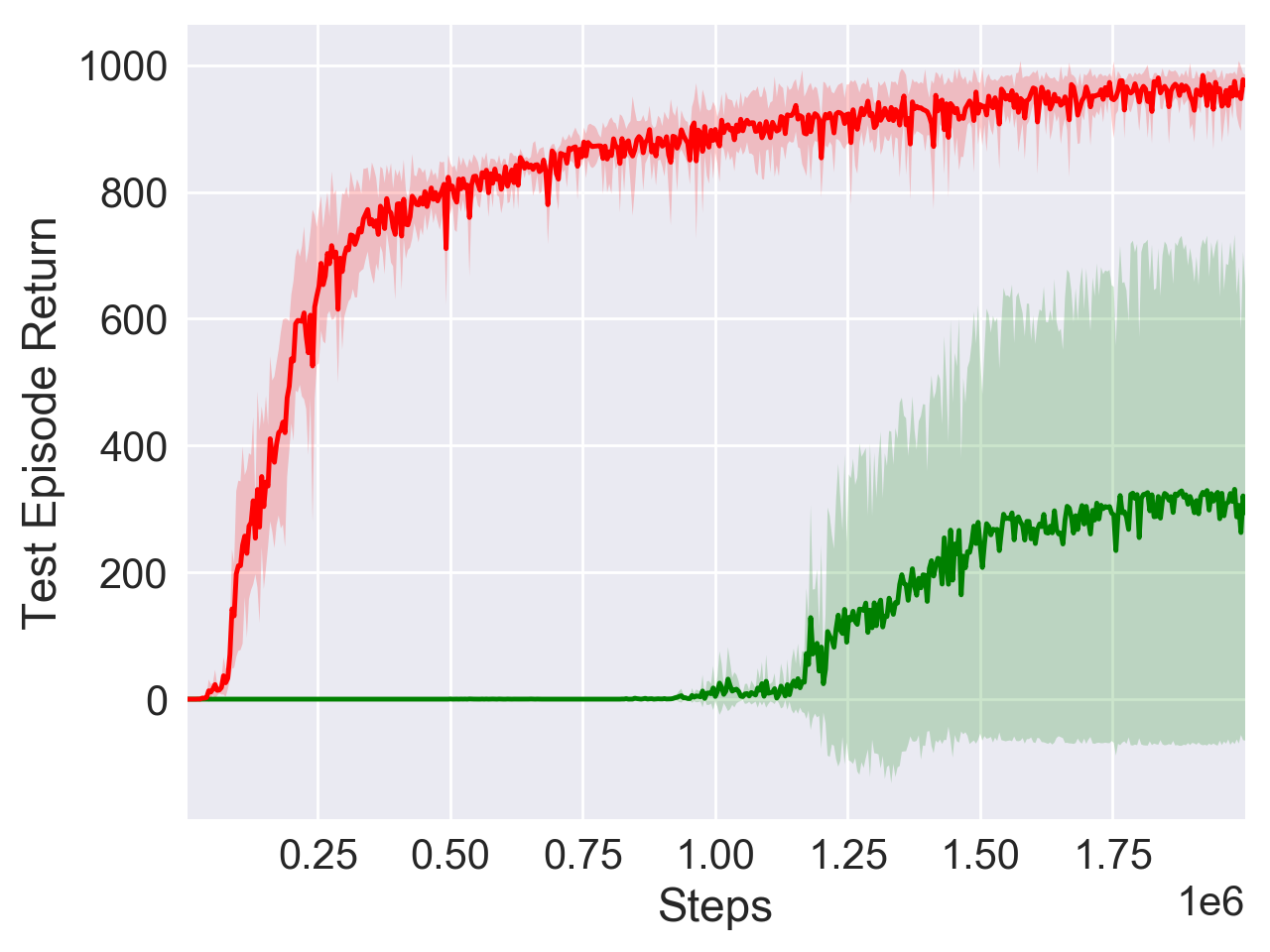}
    \end{minipage}
    \begin{minipage}[b]{.3\linewidth}
        \centering
        \subcaption{Cheetah-Run}
        \includegraphics[width=\linewidth]{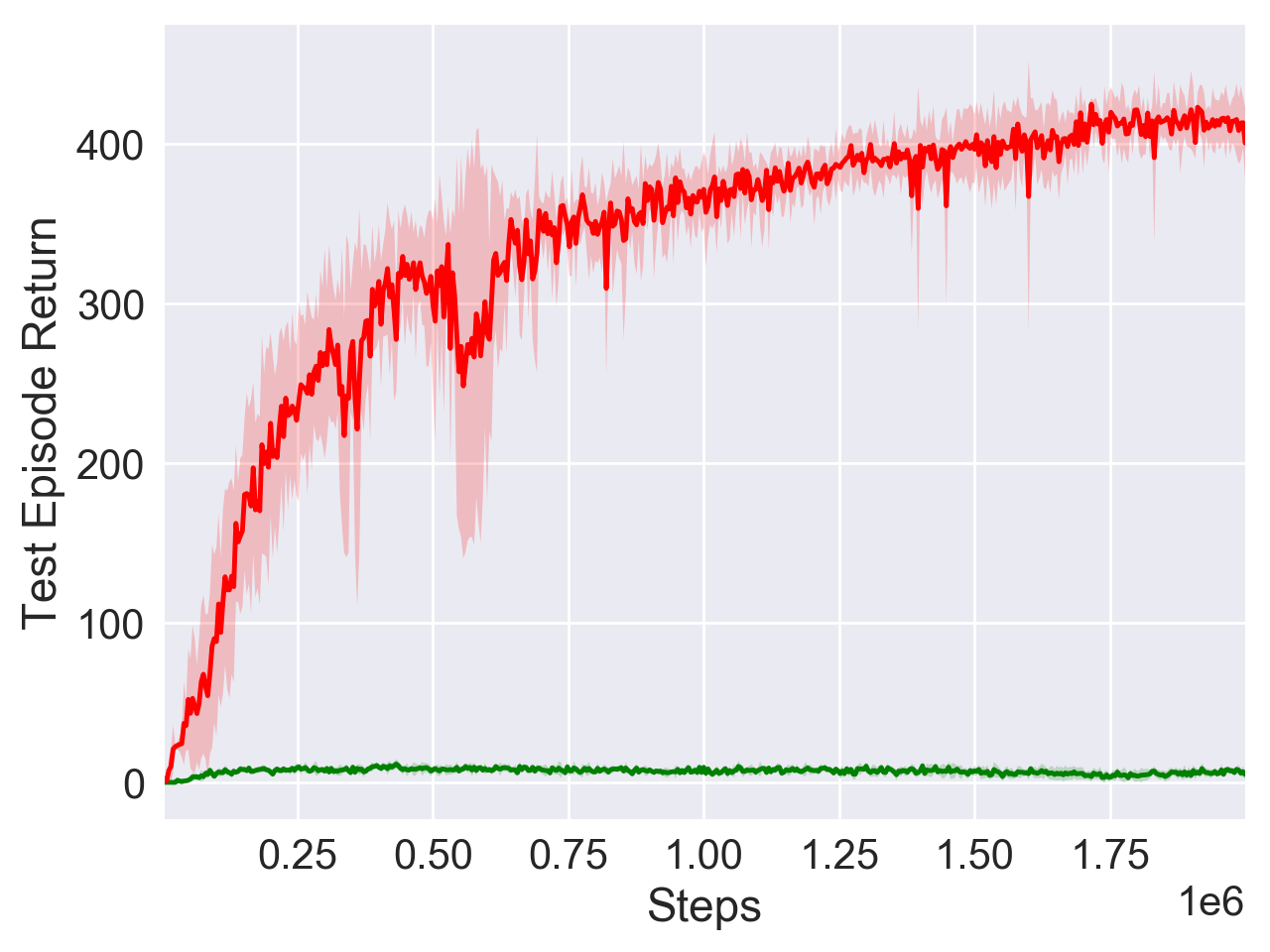}
    \end{minipage}\hfil%
    \begin{minipage}[b]{.3\linewidth}
        \centering
        \subcaption{Reacher Easy}
        \includegraphics[width=\linewidth]{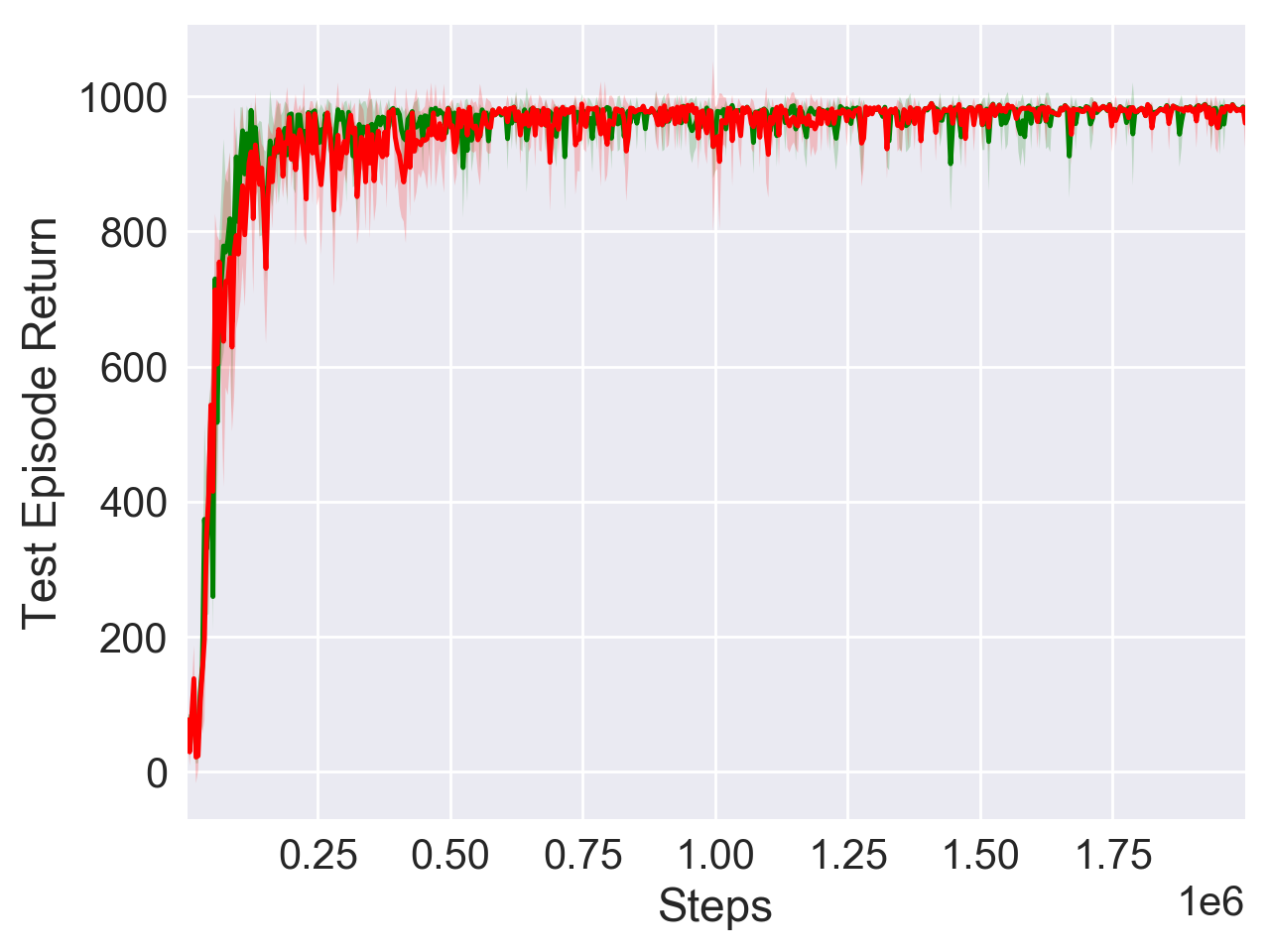}
    \end{minipage}\hfil%
    \begin{minipage}[b]{.3\linewidth}
        \centering
        \subcaption{Reacher Hard}
        \includegraphics[width=\linewidth]{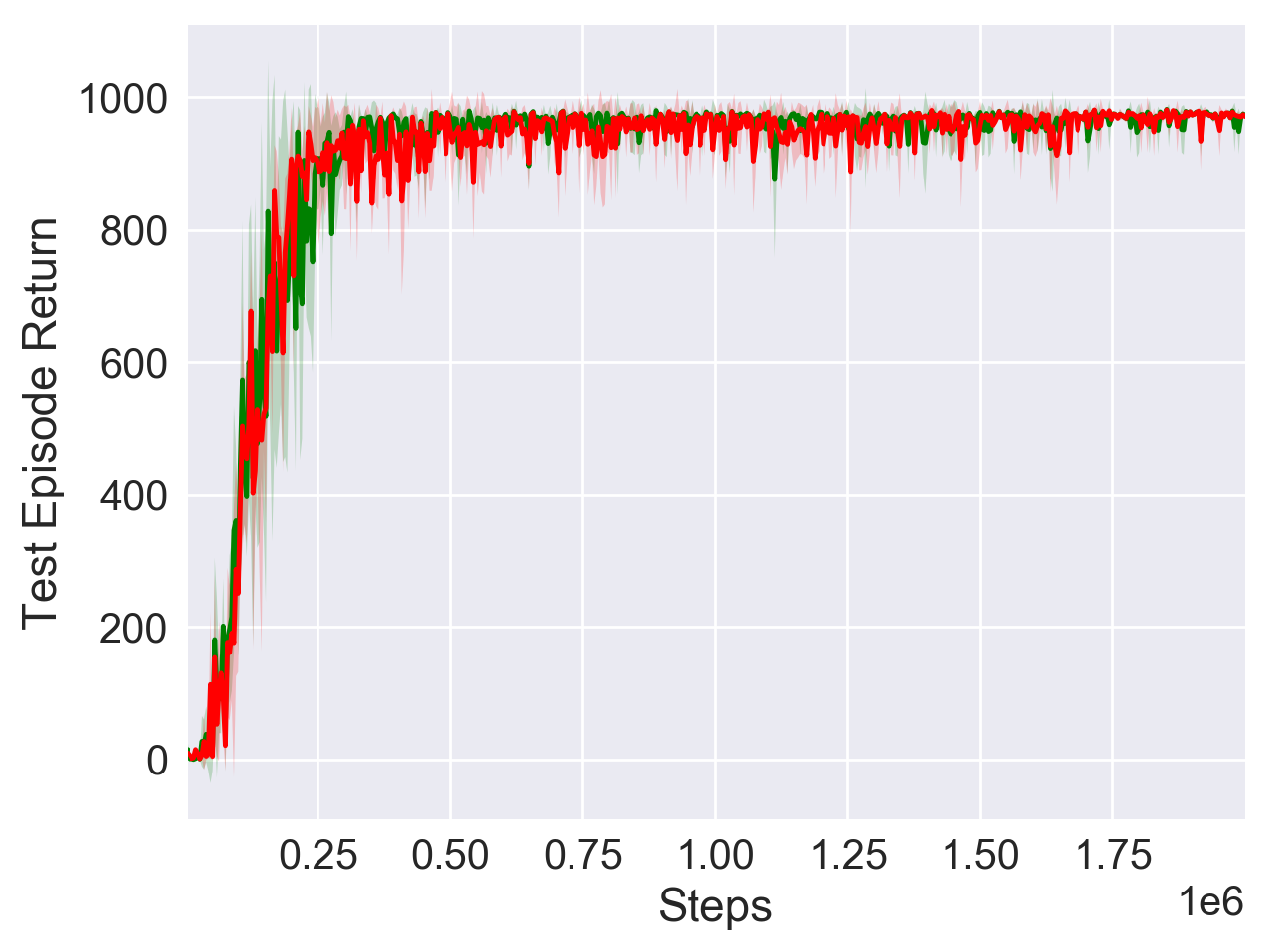}
    \end{minipage}\hfil%
    
\caption{%Results for SARC \textcolor {red}{\textit{(red)}} and SAC on a mix of PyBullet Envrironments and DeepMind Control Suite tasks after increasing Actor and Critic network sizes to [400,300]. 
Even after increasing network size to [400,300], SARC \textcolor{red}{(red)} continues to provide consistent gains over SAC. The x-axis shows the timesteps in each environment. The y-axis shows the average value and standard deviation band of return across 5 seeds (10 test episode returns per seed).}
\label{fig:network-increase}
\end{figure}

In figure~\ref{fig:network-increase}, we present the mean Monte Carlo returns over 10 test episodes at various steps during training. The x-axis shows the timesteps in each environment and the y-axis shows the mean and standard deviation band of above specified returns across 5 seeds. It can be observed that even on increasing the network sizes, SARC continues to provide consistent improvement in episode return value over SAC. The experiments as described in this section have also been performed on 5 tasks provided in PyBullet Environments \cite{coumans2016pybullet}. These can be found in Appendix B.

\section{Conclusion and Future Work}\label{section:conclusion}
%\textcolor{red}{[SARC premise]. [Properties of retrospective loss]. Extensive experimental analysis on multiple environments and achieve consistent and significant performance gain without any additional parameter tuning. As a future direction, we aim to i) extend retrospective constraints to other actor-critic algorithms and ii) generalize the retrospective loss to a parametric family of objective functions. We introduce the generalized family in definition 2.}

In this work, we presented Soft Actor Retrospective Critic (SARC). SARC builds on and improves SAC by applying a retrospective regulariser on the critic. This retrospective regulariser is inspired from retrospective loss, a recent technique that improves accuracy in the supervised setting. We proposed that using the retrospective loss as a regularizer in the critic objective accelerates critic learning. Due to the two-time scale nature of actor-critic algorithms, accelerating critic convergence can provide a better gradient estimate for the actor throughout training and improve final performance. We validated our claims of improvement in final performance through extensive experimental evaluation and analysis on multiple continuous control environments. We also compared retrospective loss with an alternate baseline technique for critic acceleration. We presented evidence of critic acceleration in SARC. We also tuned the SAC entropy regularizer, $\alpha$ to demonstrate that SARC continues providing performance gain across different values of $\alpha$. As a future direction, we aim to extend retrospective regularization to other actor-critic algorithms.

%We demonstrated that SARC significantly outperforms SAC and is competitive with other actor-critic algorithms under several experimental settings. 

\bibliographystyle{plainnat}
\bibliography{main}

\begin{thebibliography}{28}
\providecommand{\natexlab}[1]{#1}
\providecommand{\url}[1]{\texttt{#1}}
\expandafter\ifx\csname urlstyle\endcsname\relax
  \providecommand{\doi}[1]{doi: #1}\else
  \providecommand{\doi}{doi: \begingroup \urlstyle{rm}\Url}\fi

\bibitem[Achiam(2018)]{SpinningUp2018}
Joshua Achiam.
\newblock {Spinning Up in Deep Reinforcement Learning}.
\newblock 2018.

\bibitem[Bahdanau et~al.(2016)Bahdanau, Brakel, Xu, Goyal, Lowe, Pineau,
  Courville, and Bengio]{bahdanau2016actor}
Dzmitry Bahdanau, Philemon Brakel, Kelvin Xu, Anirudh Goyal, Ryan Lowe, Joelle
  Pineau, Aaron Courville, and Yoshua Bengio.
\newblock An actor-critic algorithm for sequence prediction.
\newblock \emph{arXiv preprint arXiv:1607.07086}, 2016.

\bibitem[Borkar and Konda(1997)]{borkar1997actor}
Vivek~S Borkar and Vijaymohan~R Konda.
\newblock The actor-critic algorithm as multi-time-scale stochastic
  approximation.
\newblock \emph{Sadhana}, 22\penalty0 (4):\penalty0 525--543, 1997.

\bibitem[Coumans and Bai(2016)]{coumans2016pybullet}
Erwin Coumans and Yunfei Bai.
\newblock Pybullet, a python module for physics simulation for games, robotics
  and machine learning.
\newblock 2016.

\bibitem[Fujimoto et~al.(2018)Fujimoto, Hoof, and
  Meger]{fujimoto2018addressing}
Scott Fujimoto, Herke Hoof, and David Meger.
\newblock Addressing function approximation error in actor-critic methods.
\newblock In \emph{International Conference on Machine Learning}, pages
  1587--1596. PMLR, 2018.

\bibitem[Haarnoja et~al.(2018)Haarnoja, Zhou, Abbeel, and
  Levine]{haarnoja2018soft}
Tuomas Haarnoja, Aurick Zhou, Pieter Abbeel, and Sergey Levine.
\newblock Soft actor-critic: Off-policy maximum entropy deep reinforcement
  learning with a stochastic actor.
\newblock In \emph{International Conference on Machine Learning}, pages
  1861--1870. PMLR, 2018.

\bibitem[Henderson et~al.(2018)Henderson, Islam, Bachman, Pineau, Precup, and
  Meger]{henderson2018deep}
Peter Henderson, Riashat Islam, Philip Bachman, Joelle Pineau, Doina Precup,
  and David Meger.
\newblock Deep reinforcement learning that matters.
\newblock In \emph{Proceedings of the AAAI Conference on Artificial
  Intelligence}, volume~32, 2018.

\bibitem[Jandial et~al.(2020)Jandial, Chopra, Sarkar, Gupta, Krishnamurthy, and
  Balasubramanian]{jandial2020retrospective}
Surgan Jandial, Ayush Chopra, Mausoom Sarkar, Piyush Gupta, Balaji
  Krishnamurthy, and Vineeth Balasubramanian.
\newblock Retrospective loss: Looking back to improve training of deep neural
  networks.
\newblock In \emph{Proceedings of the 26th ACM SIGKDD International Conference
  on Knowledge Discovery \& Data Mining}, pages 1123--1131, 2020.

\bibitem[Jin et~al.(2018)Jin, Allen-Zhu, Bubeck, and Jordan]{jin2018q}
Chi Jin, Zeyuan Allen-Zhu, Sebastien Bubeck, and Michael~I Jordan.
\newblock Is q-learning provably efficient?
\newblock \emph{arXiv preprint arXiv:1807.03765}, 2018.

\bibitem[Konda and Tsitsiklis(2000)]{konda2000actor}
Vijay~R Konda and John~N Tsitsiklis.
\newblock Actor-critic algorithms.
\newblock In \emph{Advances in neural information processing systems}, pages
  1008--1014. Citeseer, 2000.

\bibitem[Lillicrap et~al.(2015)Lillicrap, Hunt, Pritzel, Heess, Erez, Tassa,
  Silver, and Wierstra]{lillicrap2015continuous}
Timothy~P Lillicrap, Jonathan~J Hunt, Alexander Pritzel, Nicolas Heess, Tom
  Erez, Yuval Tassa, David Silver, and Daan Wierstra.
\newblock Continuous control with deep reinforcement learning.
\newblock \emph{arXiv preprint arXiv:1509.02971}, 2015.

\bibitem[Mnih et~al.(2013)Mnih, Kavukcuoglu, Silver, Graves, Antonoglou,
  Wierstra, and Riedmiller]{mnih2013playing}
Volodymyr Mnih, Koray Kavukcuoglu, David Silver, Alex Graves, Ioannis
  Antonoglou, Daan Wierstra, and Martin Riedmiller.
\newblock Playing atari with deep reinforcement learning.
\newblock \emph{arXiv preprint arXiv:1312.5602}, 2013.

\bibitem[Mnih et~al.(2015)Mnih, Kavukcuoglu, Silver, Rusu, Veness, Bellemare,
  Graves, Riedmiller, Fidjeland, Ostrovski, et~al.]{mnih2015human}
Volodymyr Mnih, Koray Kavukcuoglu, David Silver, Andrei~A Rusu, Joel Veness,
  Marc~G Bellemare, Alex Graves, Martin Riedmiller, Andreas~K Fidjeland, Georg
  Ostrovski, et~al.
\newblock Human-level control through deep reinforcement learning.
\newblock \emph{nature}, 518\penalty0 (7540):\penalty0 529--533, 2015.

\bibitem[Mnih et~al.(2016)Mnih, Badia, Mirza, Graves, Lillicrap, Harley,
  Silver, and Kavukcuoglu]{mnih2016asynchronous}
Volodymyr Mnih, Adria~Puigdomenech Badia, Mehdi Mirza, Alex Graves, Timothy
  Lillicrap, Tim Harley, David Silver, and Koray Kavukcuoglu.
\newblock Asynchronous methods for deep reinforcement learning.
\newblock In \emph{International conference on machine learning}, pages
  1928--1937. PMLR, 2016.

\bibitem[Schulman et~al.(2015)Schulman, Levine, Abbeel, Jordan, and
  Moritz]{schulman2015trust}
John Schulman, Sergey Levine, Pieter Abbeel, Michael Jordan, and Philipp
  Moritz.
\newblock Trust region policy optimization.
\newblock In \emph{International conference on machine learning}, pages
  1889--1897. PMLR, 2015.

\bibitem[Schulman et~al.(2017)Schulman, Wolski, Dhariwal, Radford, and
  Klimov]{schulman2017proximal}
John Schulman, Filip Wolski, Prafulla Dhariwal, Alec Radford, and Oleg Klimov.
\newblock Proximal policy optimization algorithms.
\newblock \emph{arXiv preprint arXiv:1707.06347}, 2017.

\bibitem[Silver et~al.(2014)Silver, Lever, Heess, Degris, Wierstra, and
  Riedmiller]{silver2014deterministic}
David Silver, Guy Lever, Nicolas Heess, Thomas Degris, Daan Wierstra, and
  Martin Riedmiller.
\newblock Deterministic policy gradient algorithms.
\newblock In \emph{International conference on machine learning}, pages
  387--395. PMLR, 2014.

\bibitem[Sutton and Barto(2018)]{sutton2018reinforcement}
Richard~S Sutton and Andrew~G Barto.
\newblock \emph{Reinforcement learning: An introduction}.
\newblock MIT press, 2018.

\bibitem[Sutton et~al.(1999)Sutton, McAllester, Singh, Mansour,
  et~al.]{sutton1999policy}
Richard~S Sutton, David~A McAllester, Satinder~P Singh, Yishay Mansour, et~al.
\newblock Policy gradient methods for reinforcement learning with function
  approximation.
\newblock In \emph{NIPs}, volume~99, pages 1057--1063. Citeseer, 1999.

\bibitem[Tassa et~al.(2018)Tassa, Doron, Muldal, Erez, Li, Casas, Budden,
  Abdolmaleki, Merel, Lefrancq, et~al.]{tassa2018deepmind}
Yuval Tassa, Yotam Doron, Alistair Muldal, Tom Erez, Yazhe Li, Diego de~Las
  Casas, David Budden, Abbas Abdolmaleki, Josh Merel, Andrew Lefrancq, et~al.
\newblock Deepmind control suite.
\newblock \emph{arXiv preprint arXiv:1801.00690}, 2018.

\bibitem[Thrun and Schwartz(1993)]{thrun1993issues}
Sebastian Thrun and Anton Schwartz.
\newblock Issues in using function approximation for reinforcement learning.
\newblock In \emph{Proceedings of the Fourth Connectionist Models Summer
  School}, pages 255--263. Hillsdale, NJ, 1993.

\bibitem[{Todorov} et~al.(2012){Todorov}, {Erez}, and {Tassa}]{6386109}
E.~{Todorov}, T.~{Erez}, and Y.~{Tassa}.
\newblock Mujoco: A physics engine for model-based control.
\newblock In \emph{2012 IEEE/RSJ International Conference on Intelligent Robots
  and Systems}, pages 5026--5033, 2012.
\newblock \doi{10.1109/IROS.2012.6386109}.

\bibitem[Van~Hasselt et~al.(2016)Van~Hasselt, Guez, and Silver]{van2016deep}
Hado Van~Hasselt, Arthur Guez, and David Silver.
\newblock Deep reinforcement learning with double q-learning.
\newblock In \emph{Proceedings of the AAAI Conference on Artificial
  Intelligence}, volume~30, 2016.

\bibitem[Wang et~al.(2016)Wang, Bapst, Heess, Mnih, Munos, Kavukcuoglu, and
  de~Freitas]{wang2016sample}
Ziyu Wang, Victor Bapst, Nicolas Heess, Volodymyr Mnih, Remi Munos, Koray
  Kavukcuoglu, and Nando de~Freitas.
\newblock Sample efficient actor-critic with experience replay.
\newblock \emph{arXiv preprint arXiv:1611.01224}, 2016.

\bibitem[Wiering(2004)]{wiering2004convergence}
Marco~A Wiering.
\newblock Convergence and divergence in standard and averaging reinforcement
  learning.
\newblock In \emph{European Conference on Machine Learning}, pages 477--488.
  Springer, 2004.

\bibitem[Williams(1992)]{williams1992simple}
Ronald~J Williams.
\newblock Simple statistical gradient-following algorithms for connectionist
  reinforcement learning.
\newblock \emph{Machine learning}, 8\penalty0 (3-4):\penalty0 229--256, 1992.

\bibitem[Wu et~al.(2020)Wu, Zhang, Xu, and Gu]{wu2020finite}
Yue Wu, Weitong Zhang, Pan Xu, and Quanquan Gu.
\newblock A finite time analysis of two time-scale actor critic methods.
\newblock \emph{arXiv preprint arXiv:2005.01350}, 2020.

\bibitem[Wu et~al.(2017)Wu, Mansimov, Liao, Grosse, and Ba]{wu2017scalable}
Yuhuai Wu, Elman Mansimov, Shun Liao, Roger Grosse, and Jimmy Ba.
\newblock Scalable trust-region method for deep reinforcement learning using
  kronecker-factored approximation.
\newblock \emph{arXiv preprint arXiv:1708.05144}, 2017.

\end{thebibliography}

\clearpage

\appendix

\section{PyBullet Results for Section 5: Results}

\begin{figure*}[b]
  \centering
  \begin{minipage}[b]{.38\linewidth}
    \centering
    \captionof{table}{Test episode returns of SARC and SAC on 5 PyBullet Environments after training for 2M steps.}
     \label{table:spinningup-pybullet}
     \centering
     \footnotesize
     \begin{tabular}{lrr}
     \toprule
         \multirow{2}{*}{\textbf{Environment}} & \multicolumn{2}{c}{\textbf{Avg Test Ep Return}}\\ \cmidrule(r){2-3}
        
         & \textbf{SAC} & \textbf{SARC}\\
          \midrule
        
        Ant & 1581.59 & \textbf{2333.26} \\
        Half Cheetah & 1701.19 & \textbf{2734.0}\\
        Hopper & \textbf{2497.83} & 1975.58\\
      Reacher & \textbf{20.55} & \textbf{20.05}\\
        Walked-2D & 1093.64 & \textbf{1761.26}\\
        
         \bottomrule
     \end{tabular}
     \vspace{0pt}
    \end{minipage}\hfil%
    \begin{minipage}[b]{.28\linewidth}
        \centering
        \subcaption{Hopper}
        \includegraphics[width=\linewidth]{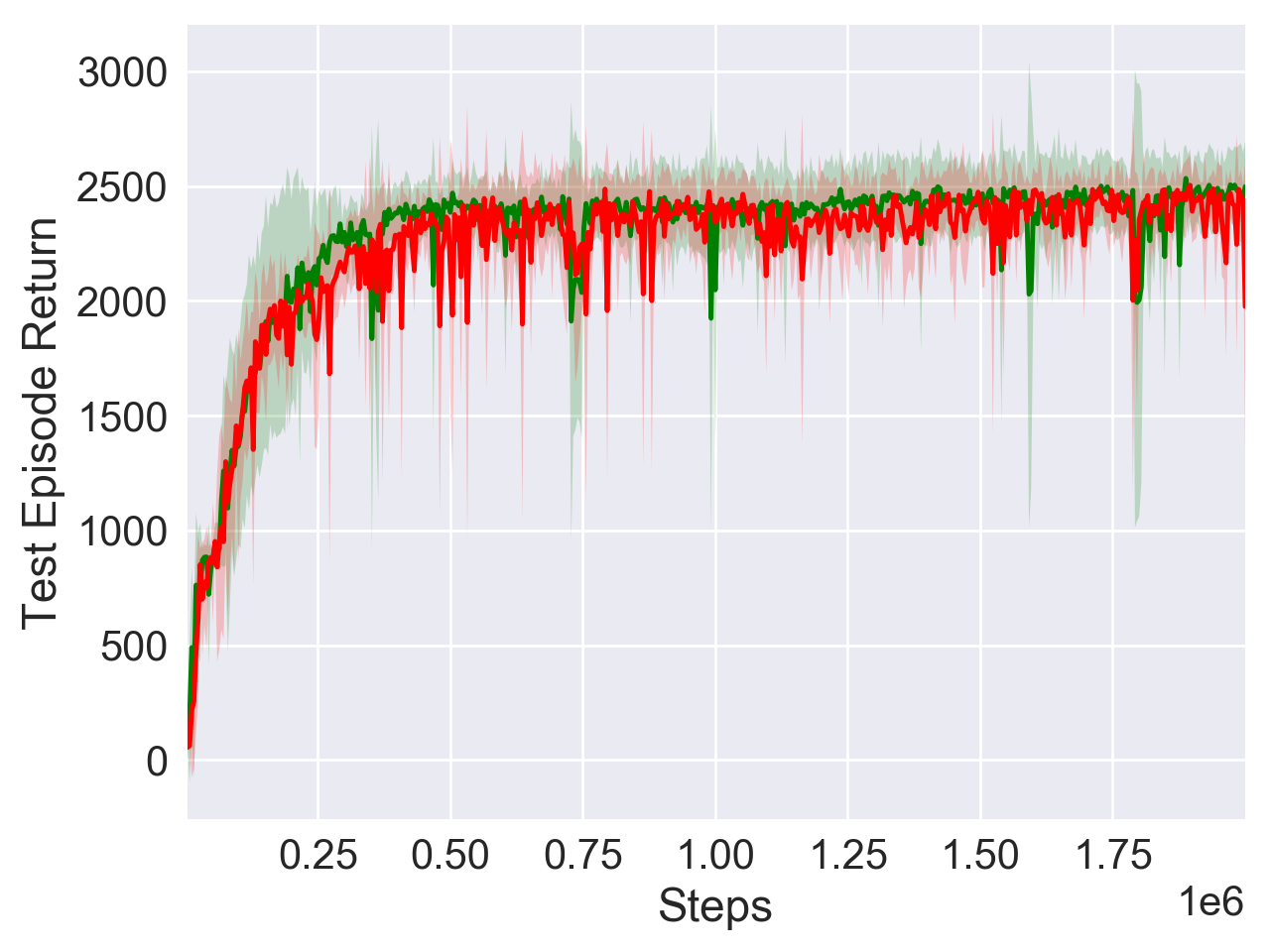}
    \end{minipage}\hfil%
    \begin{minipage}[b]{.28\linewidth}
        \centering
        \subcaption{Reacher}
        \includegraphics[width=\linewidth]{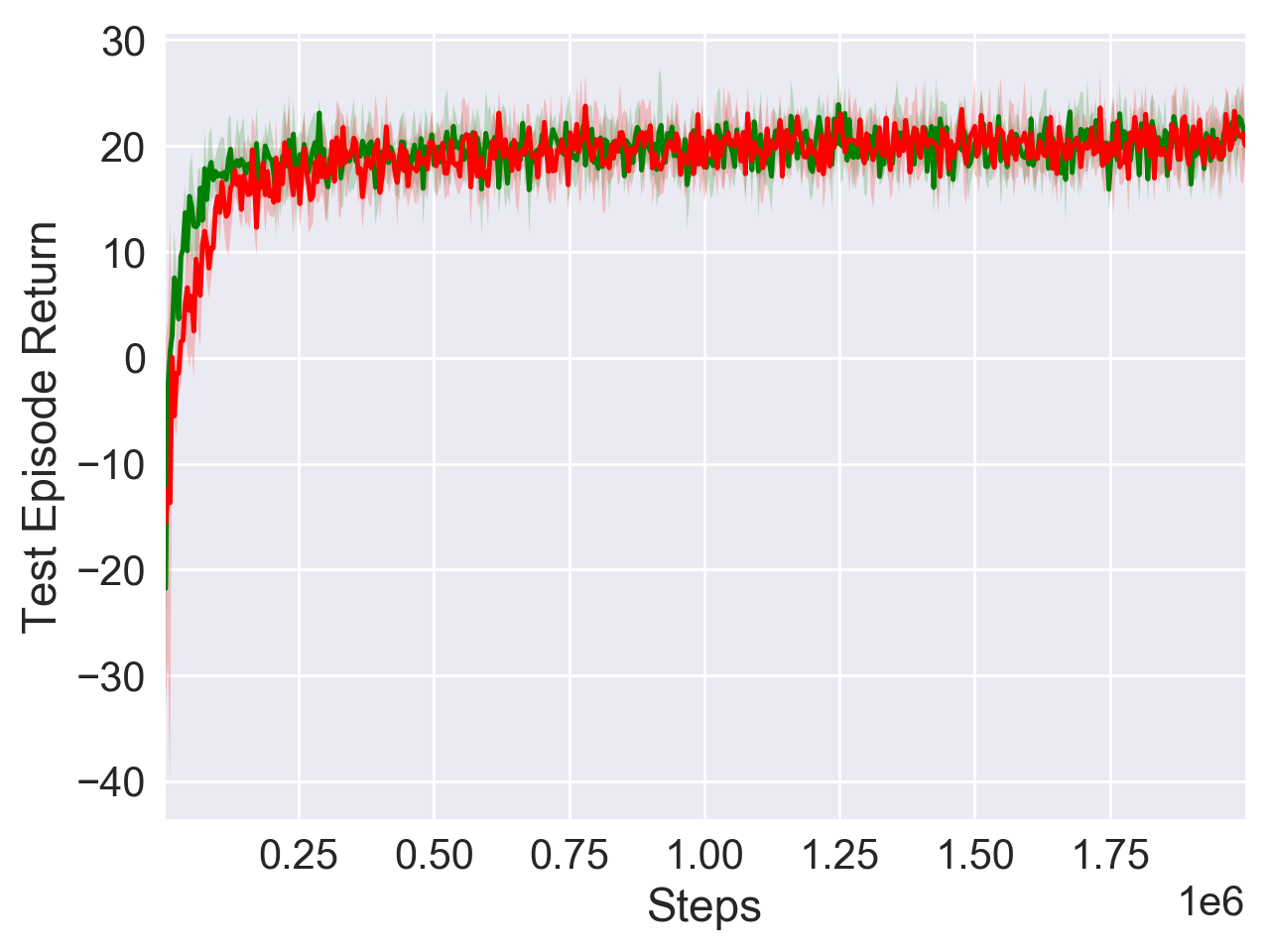}
    \end{minipage}\hfil%
    \begin{minipage}[b]{.32\linewidth}
        \centering
        \subcaption{Ant}
        \includegraphics[width=\linewidth]{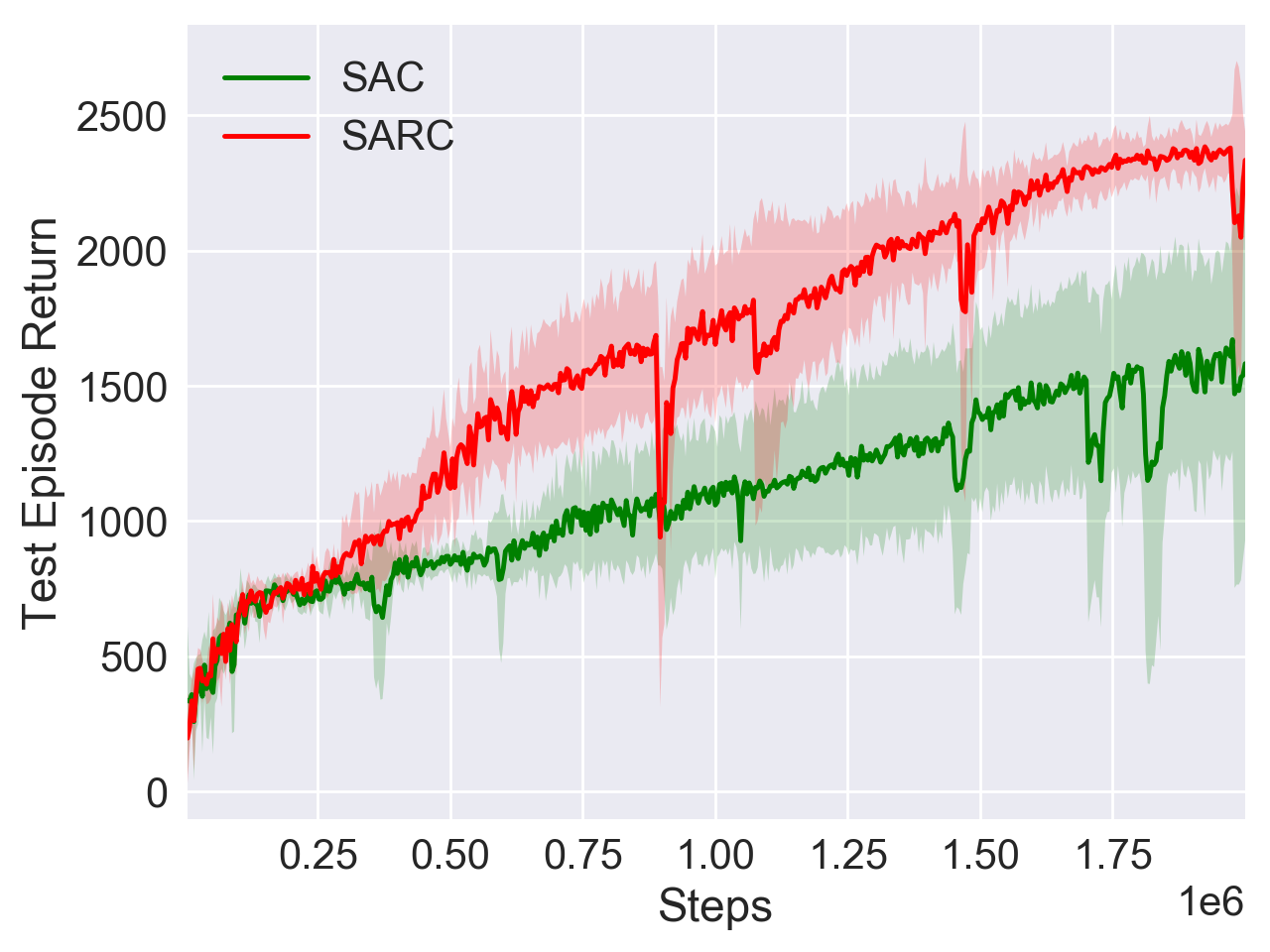}
    \end{minipage}\hfil%
    \begin{minipage}[b]{.32\linewidth}
        \centering
        \subcaption{Half Cheetah}
        \includegraphics[width=\linewidth]{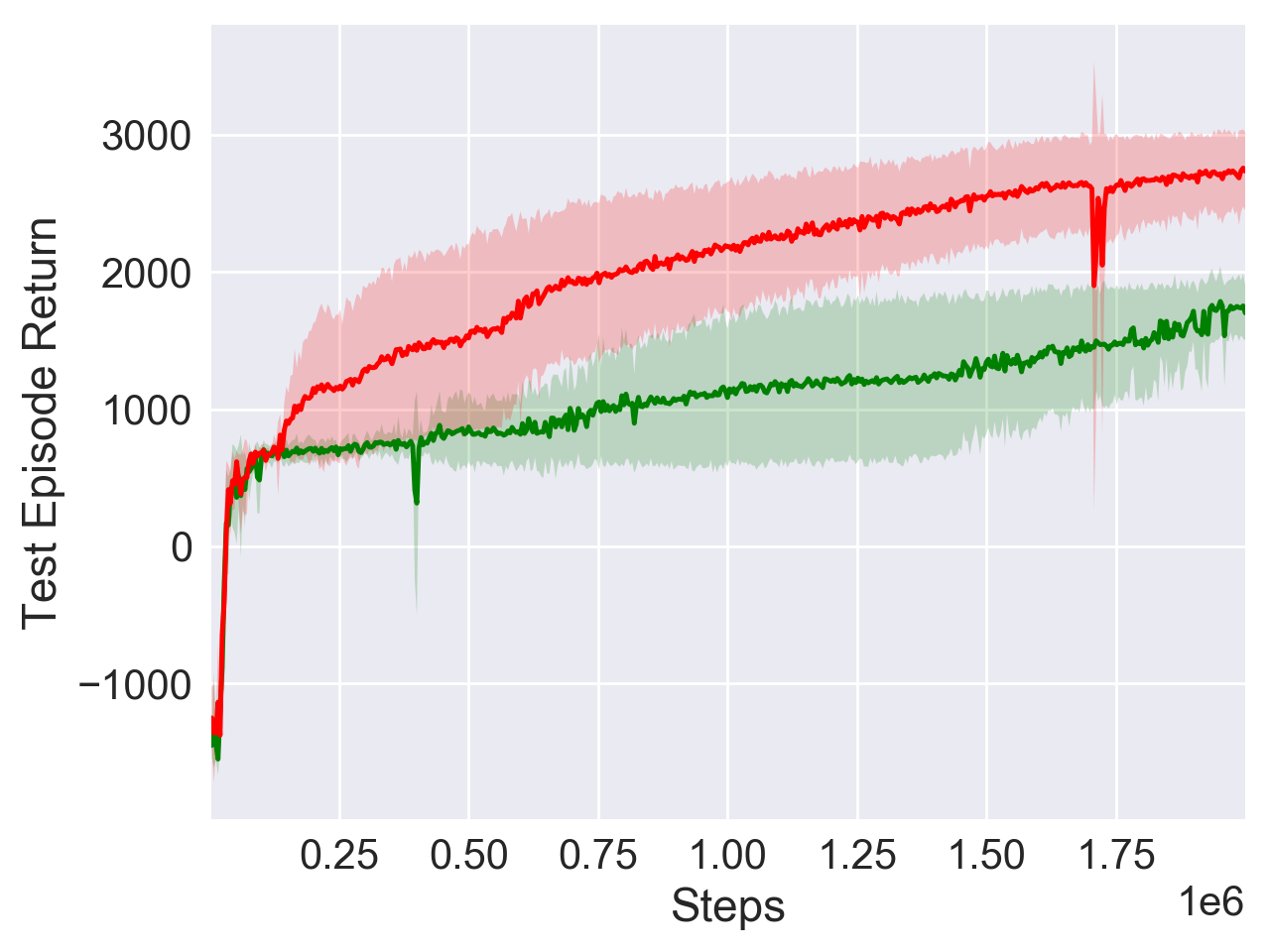}
    \end{minipage}\hfil%
    \begin{minipage}[b]{.32\linewidth}
        \centering
        \subcaption{Walker-2D}
        \includegraphics[width=\linewidth]{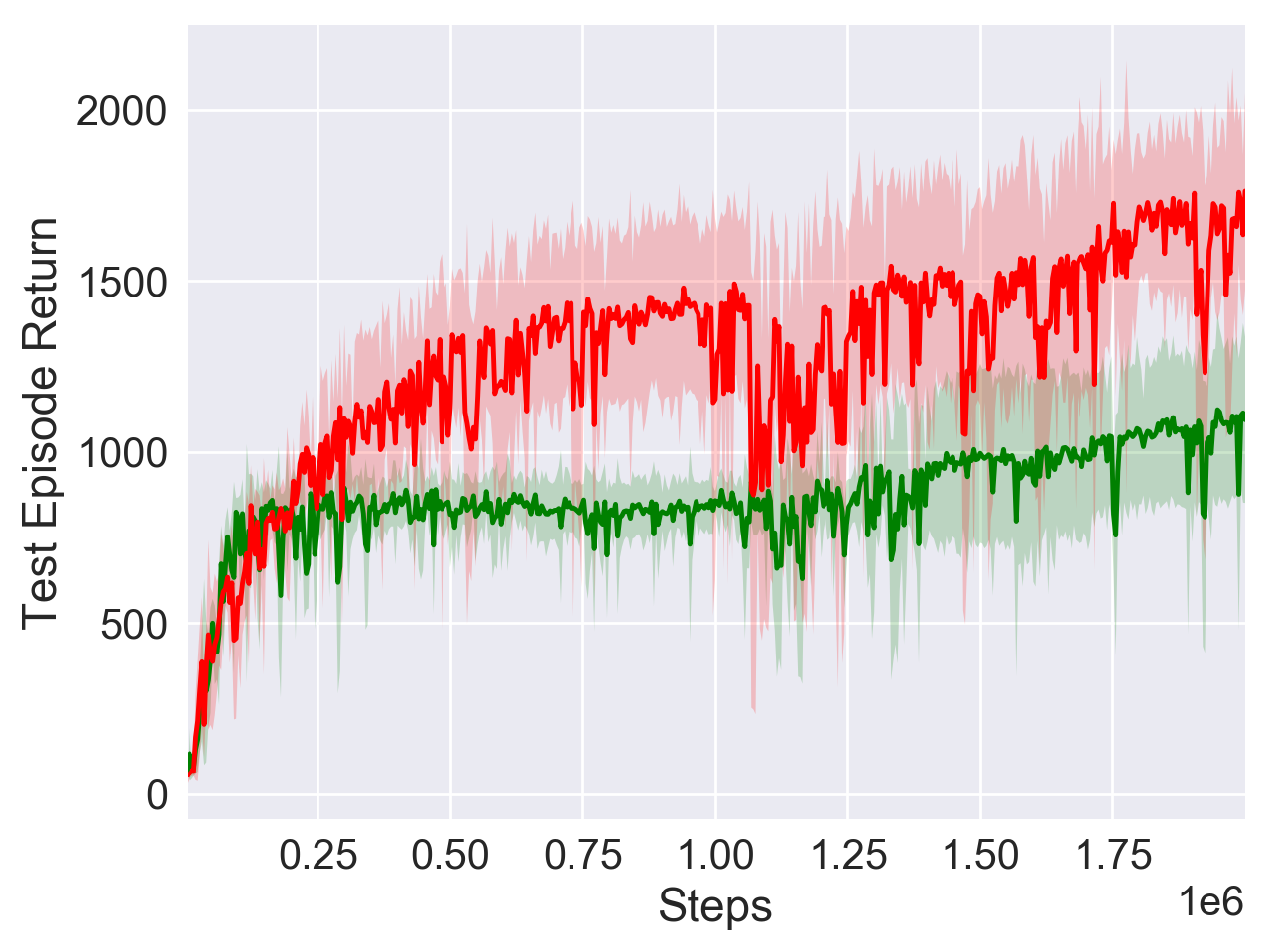}
    \end{minipage}%
\caption{Results for SARC \textcolor{red}{(red)} and SAC on 5 PyBullet Environments. The x-axis shows the timesteps in each environment. The y-axis shows the average value and standard deviation band of return across 5 seeds (10 test episode returns per seed). It can be observed that SARC improves SAC in most environments even after modifying the underlying physics engine.}
\label{fig:spinningup-pybullet}
\end{figure*}

We have demonstrated SARC using tasks from the DeepMind Control Suite which uses the MuJoCo Physics Engine \cite{6386109}. In this section, we present comparisons on a set of environments that use the Bullet Physics Engine \cite{coumans2016pybullet}, an open-source physics simulator. We perform these evaluations on 5 PyBullet \cite{coumans2016pybullet} environments: Ant, HalfCheetah, Hopper, Reacher, Walker-2D.

Table \ref{table:spinningup-pybullet} shows the average value across 5 seeds (10 test episode returns per seed) obtained for each of the 5 tasks after training for 2 million timesteps. It can be observed that, at the end of training, SARC outperforms or is at par with existing approaches on \textbf{4 out of 5 environments} presented in \textbf{bold}. 

In figure~\ref{fig:spinningup-pybullet}, we present the mean Monte Carlo returns over 10 test episodes at various steps during training. The x-axis shows the timesteps in each environment and the y-axis shows the mean and standard deviation band of above specified returns across 5 seeds.

It can be observed that, on Ant, Half-Cheetah and Walker-2D, SARC achieves a higher return value faster than SAC. On Hopper and Reacher, SARC overlaps with and does not degrade the performance of SAC. Even on changing the underlying physics simulator, SARC continues providing consistent improvement over SAC. 

We also hope that using an open-source physics simulator will help with reproducibility and reach of our work.

\begin{figure*}
   \centering
    \begin{minipage}[b]{.3\linewidth}
        \centering
        \subcaption{AntBulletEnv-v0}
        \includegraphics[width=\linewidth]{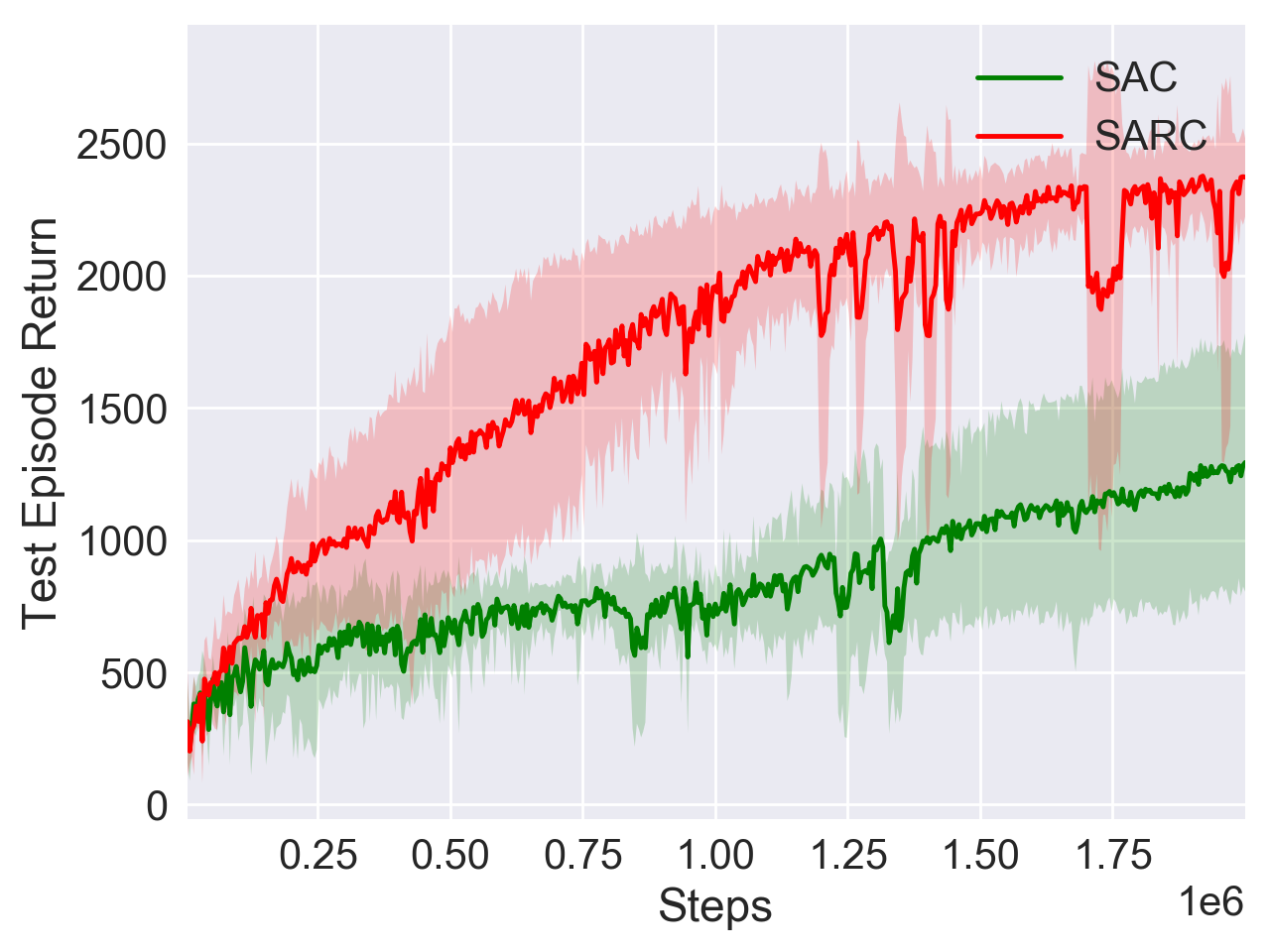}
    \end{minipage}\hfil%
    \begin{minipage}[b]{.3\linewidth}
        \centering
        \subcaption{HalfCheetahBulletEnv-v0}
        \includegraphics[width=\linewidth]{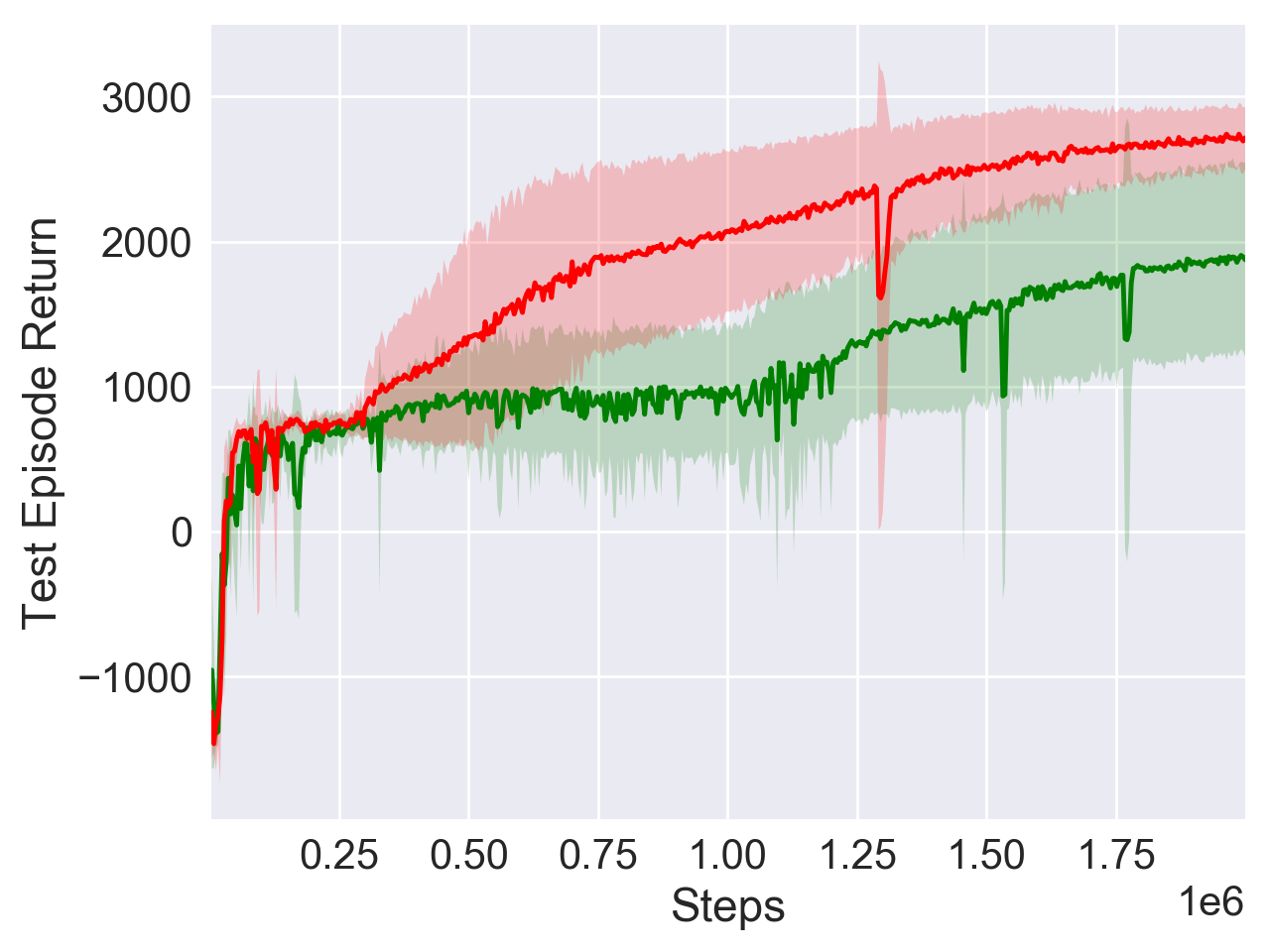}
    \end{minipage}\hfil%
    \begin{minipage}[b]{.3\linewidth}
        \centering
        \subcaption{HopperBulletEnv-v0}
        \includegraphics[width=\linewidth]{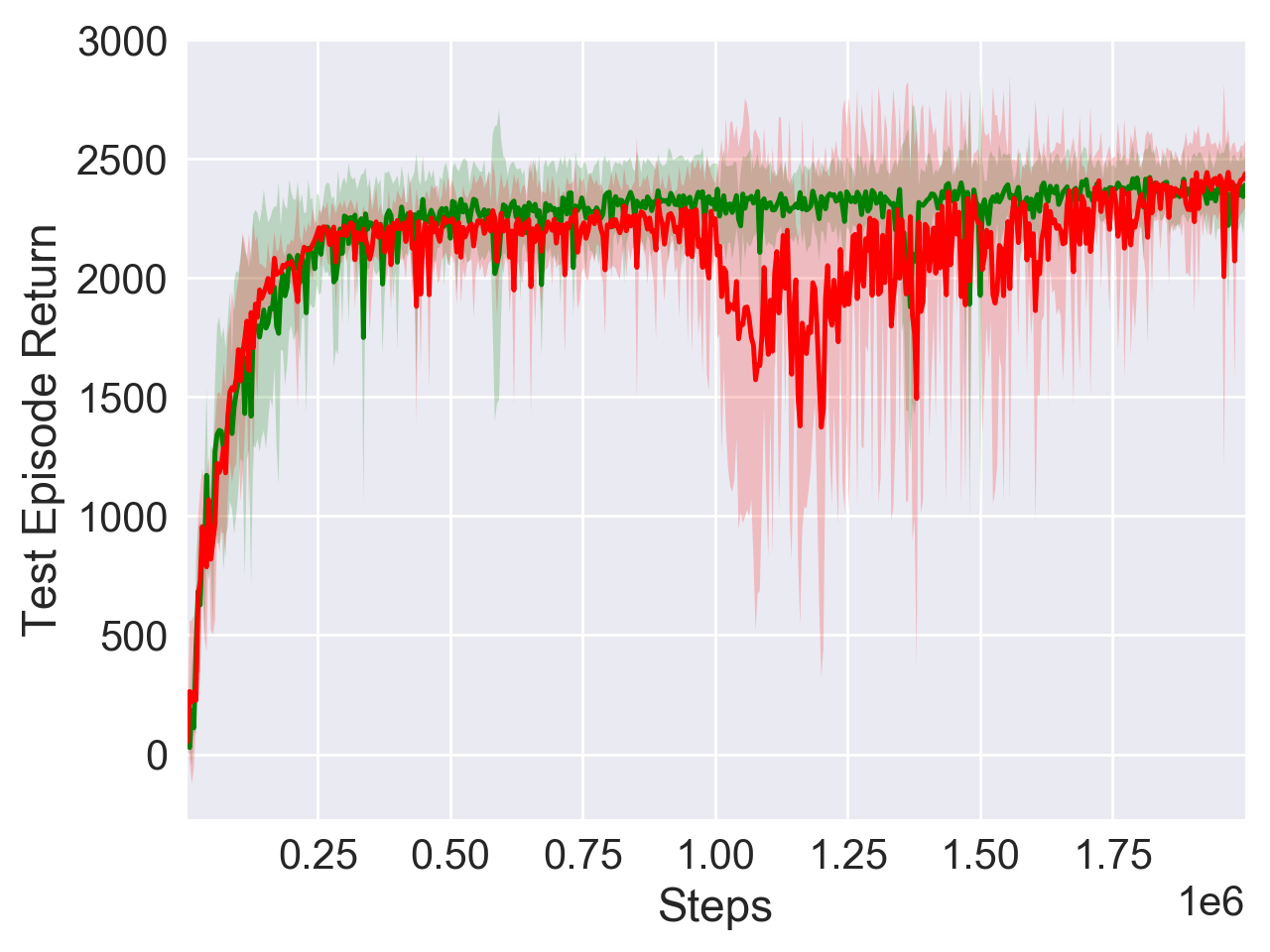}
    \end{minipage}\hfil%
    \begin{minipage}[b]{.3\linewidth}
        \centering
        \subcaption{ReacherBulletEnv-v0}
        \includegraphics[width=\linewidth]{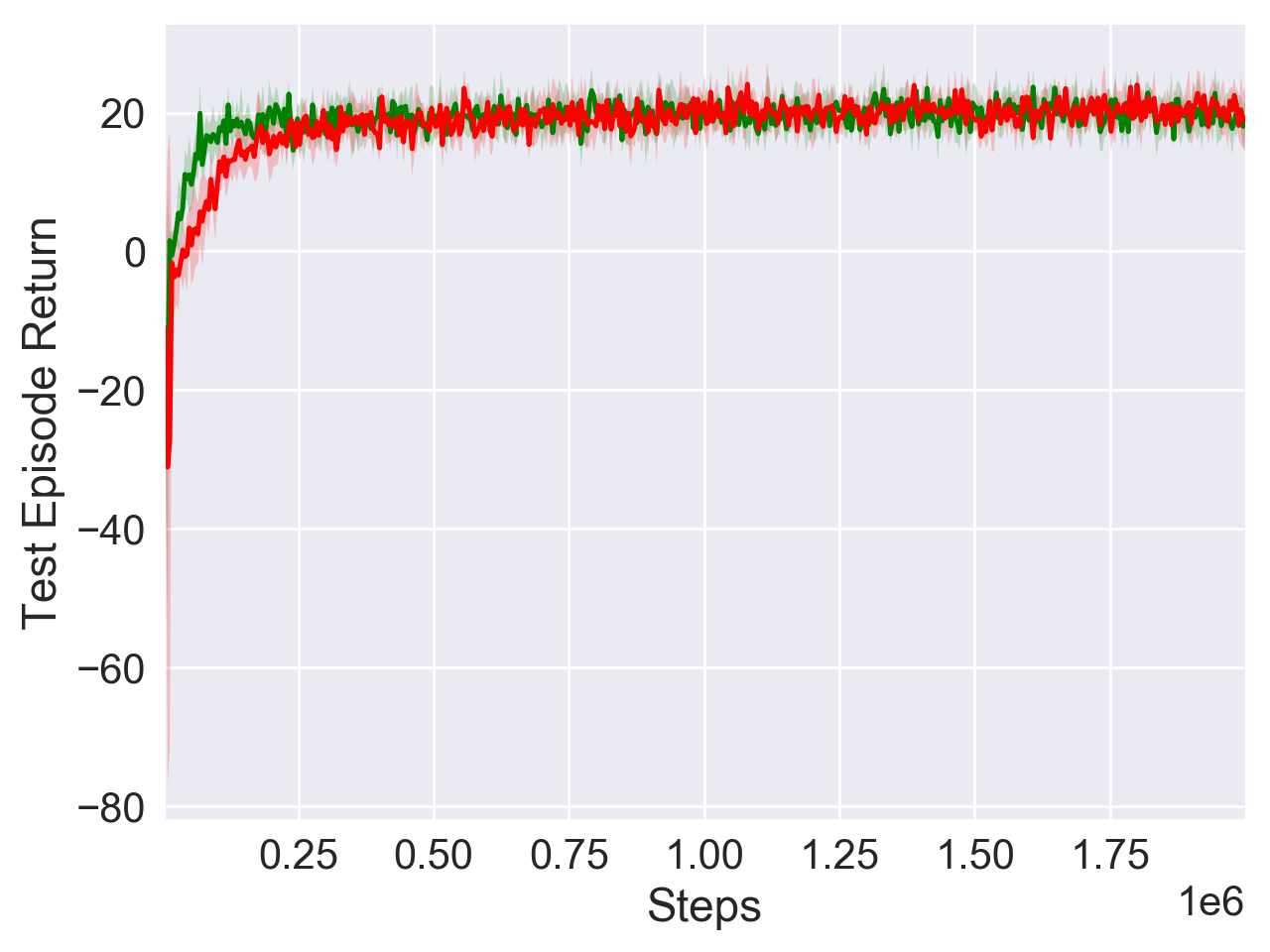}
    \end{minipage}\hfil%
    \begin{minipage}[b]{.3\linewidth}
        \centering
        \subcaption{Walker2DBulletEnv-v0}
        \includegraphics[width=\linewidth]{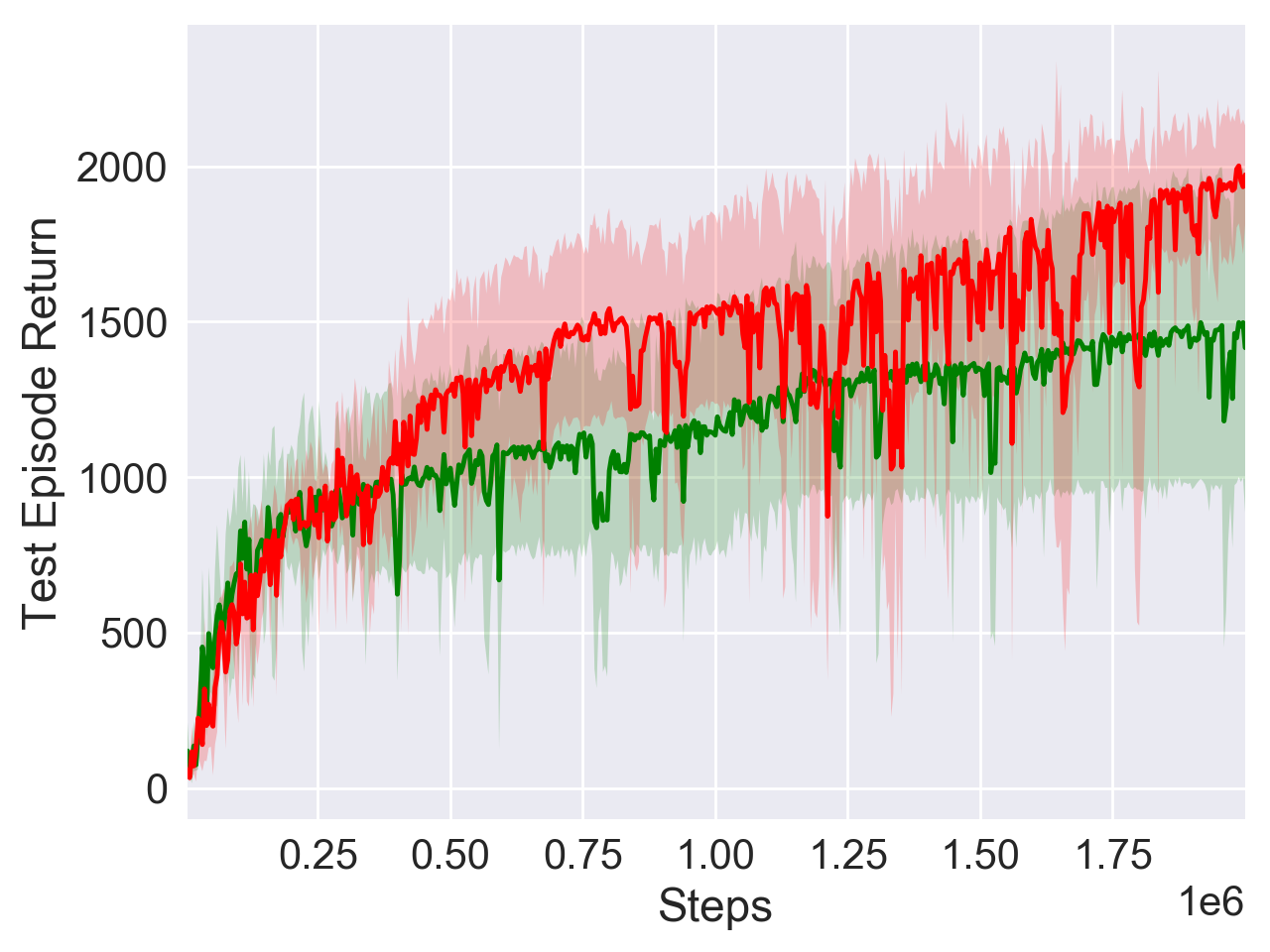}
    \end{minipage}%
\caption{Results for SARC \textcolor {red}{\textit{(red)}} and SAC on PyBullet Envrironments after increasing Actor and Critic network sizes to [400,300]. Even after increasing network size to [400,300], SARC \textcolor{red}{(red)} continues to provide consistent gains over SAC. The x-axis shows the timesteps in each environment. The y-axis shows the average value and standard deviation band of return across 5 seeds (10 test episode returns per seed)
}
\label{fig:network-increase2}
\end{figure*}

\section{PyBullet Results for Section 6.4: Increasing actor and critic network complexity}

Figure \ref{fig:network-increase2} shows results for SARC and SAC on PyBullet Envrironments after increasing Actor and Critic network sizes to [400,300]. Even after increasing network size to [400,300], SARC continues to provide consistent gains over SAC.

\end{document}